\documentclass{article} 
\usepackage{iclr2027_conference,times}

\usepackage{amsmath,amsfonts,bm}

\def\eqref#1{equation~\ref{#1}}

\def\1{\bm{1}}

\DeclareMathAlphabet{\mathsfit}{\encodingdefault}{\sfdefault}{m}{sl}
\SetMathAlphabet{\mathsfit}{bold}{\encodingdefault}{\sfdefault}{bx}{n}

\usepackage{hyperref}
\usepackage{nicematrix}
\usepackage{url}
\usepackage{hyperref}
\usepackage{url}
\usepackage{graphicx}
\usepackage{placeins}
\usepackage{amssymb}
\usepackage{wrapfig}
\usepackage{nicefrac}
\usepackage{booktabs}
\usepackage{multirow}
\usepackage{anyfontsize}
\usepackage{wrapfig}
\usepackage{algorithm, algpseudocode, amsmath, amssymb, xcolor}
\usepackage{listings}
\usepackage{subcaption} 
\usepackage{enumitem}
\usepackage{amsmath,amssymb}
\usepackage{xcolor}
\usepackage{tcolorbox}
\definecolor{algobg}{RGB}{243, 246, 252}
\definecolor{commentgray}{RGB}{140, 140, 140}
\definecolor{headerblue}{RGB}{0, 45, 160}

\title{Kinematic MeanFlow: One-Step Action Generation Policy for Robotic Foundation Models}

\iclrpreprint
\author{Jiawei Fan$^1$, Sifeng Wang$^1$, Yuqing Hou$^2$, Anbang Yao$^1$\thanks{ Corresponding author.} \\
$^1$Intel Labs China  \ \ $^2$Midea AI Research \\
\texttt{\{jiawei.fan, anbang.yao\}@intel.com} \\
}

\begin{document}

\maketitle

\begin{abstract}
In this paper, we study how to achieve one-step action generation in Robotic Foundation Models (RFMs), aiming to overcome the high inference latency of multi-step flow matching. MeanFlow provides a promising framework for this goal, yet its direct application leads to performance collapse.
We discover that this stems from two distinctive dynamics exhibited in the RFM velocity field: (1) the ``local acceleration" exhibits stability early on, but surges sharply towards the end of the denoising process, and (2) the spread of its magnitudes across samples widens as denoising progresses. 
To address these issues, we introduce \textbf{Kinematic MeanFlow (K-MF)}, a novel one-step action policy tailored for RFMs.
Specifically, grounded in a kinematic identity, K-MF decouples the time derivative term in the MeanFlow formulation into two sub-interval terms separated by an intermediate point. 
This decoupled formulation enables the two terms to capture early-stage and late-stage denoising dynamics, respectively, 
while mitigating the error amplification across the process.
As a result, our K-MF empowers RFMs to achieve one-step action generation in both training from scratch and fine-tuning paradigms across diverse tasks, while outperforming multi-step flow matching in most settings. 
In terms of inference efficiency, K-MF reduces action-head latency of GR00T-N1.6 by 67.5\%–74.4\% across L40 and Jetson Orin in eager and compiled modes, yielding end-to-end latency reductions of 30.3\%–54.9\%. Code will be available at \texttt{https://github.com/IntelChina-AI/K-MF}.

\end{abstract}

\section{Introduction}
\label{sec:intro}

\textbf{Background.} Robotic foundation models (RFMs) aim to establish a generalist robot manipulation policy
capable of real-time execution across diverse robotic embodiments.
Driven by this vision, the architecture of RFMs has continuously evolved.
The first evolution concerns the multimodal backbone. Early RFMs~\citep{jang2022bc,brohan2023rt} employ FiLM-conditioned convolutional neural networks to fuse the language instructions and visual observations.
Spurred by the remarkable generalization capabilities of large/vision language models (LLMs/VLMs)~\citep{touvron2023llama, touvron2023llama2,beyer2024paligemma}, some studies~\citep{zitkovich2023rt, kim2025openvla} leverage them as the core foundation for RFMs. Generally, such architectures integrate a pre-trained VLM\footnote{For brevity, we use ``VLM" to represent architectures that either employ a pre-trained VLM or pair a pre-trained LLM with a vision encoder, as their training and inference pipelines in RFMs are conceptually identical.} as the backbone to autoregressively generate discrete action tokens.  However, this creates a severe inference burden for real-time control and thus drives the second major evolution in action generation policies.
Recent RFMs~\citep{black2025pi0, bjorck2025gr00t,luo2026simvla} adopt transformer-based flow matching policies. Specifically, flow matching establishes a velocity field mapping Gaussian noise to valid actions. 
Using features obtained from a single VLM forward pass, the policy iteratively queries the transformer to predict \textit{instantaneous velocities} across denoising timesteps, progressively transforming noise into multiple actions in parallel.
Even as recent architectures~\citep{ye2026world, yuan2026fast} explore replacing VLM backbones with diffusion transformers (DiTs), 
flow matching remains the prevalent choice for action generation in RFMs.

\textbf{Motivation.} Although this policy enables parallel action generation, its iterative forward passes required for denoising incur non-negligible inference overhead. Taking GR00T-N1.6 as an example, action head latency occupies about 40\% of the end-to-end (E2E) inference time on desktop GPUs (e.g., NVIDIA L40) and up to 50\% on robotic edge platforms (e.g., NVIDIA Jetson Orin).
In other words, the computational burden of the action head critically reduces the overall decision frequency of RFMs.
Therefore, reducing the number of denoising steps becomes a critical issue.
From a broader perspective, MeanFlow has recently emerged in image generation as a remarkably simple alternative to complex distillation or consistency constraints for one-step generation. Unlike flow matching that predicts \textit{instantaneous velocity} $\mathbf{v}_t$ at timestep $t$, MeanFlow aims to learn the \textit{average velocity} $\mathbf{u}_{t,r}$\footnote{Here, $t,r\in[0,1]$, with $0$ denoting clean data
and $1$ denoting pure noise. We use $\mathbf{u}_{t,r}$ as shorthand
for $\mathbf{u}(\mathbf{z}_t,r,t)$, formally introduced in
Section~\ref{sec:method}. Throughout this paper, we adopt
the subscript order $(t,r)$ to reflect the denoising
direction from $t$ to $r$, although the average velocity
is defined from $r$ to $t$.} between any two timesteps $t$ and $r$.
Once trained, by directly predicting the average velocity across the endpoint timesteps, the model achieves one-step denoising in a single forward pass. In action generation, there have been a few efforts that built upon the MeanFlow formulation.
In imitation learning, MP1~\citep{sheng2026mp1} couples 3D inputs with a dispersive loss on latent features to prevent representation collapse. Other studies improve MeanFlow in reinforcement learning (RL). 
MVP~\citep{zhan2026mean} introduces an instantaneous velocity constraint to stabilize RL training; MeanflowQL~\citep{wang2026one} reformulates MeanFlow into a residual mapping to enable joint optimization with Q-learning; DMPO~\citep{zou2026one} combines dispersive regularization with PPO fine-tuning for one-step policy learning. These efforts remain focused on \textit{specialist policies}, where directly applying MeanFlow typically yields viable, albeit degraded, performance. In contrast, we empirically find that MeanFlow breaks down under the standard training recipe in \textit{generalist RFMs}, motivating us to investigate the fundamental issues underlying this failure.

\begin{figure}[]
    \centering
    \begin{subfigure}[b]{0.42\linewidth}
        \centering
        \includegraphics[width=0.75\linewidth]{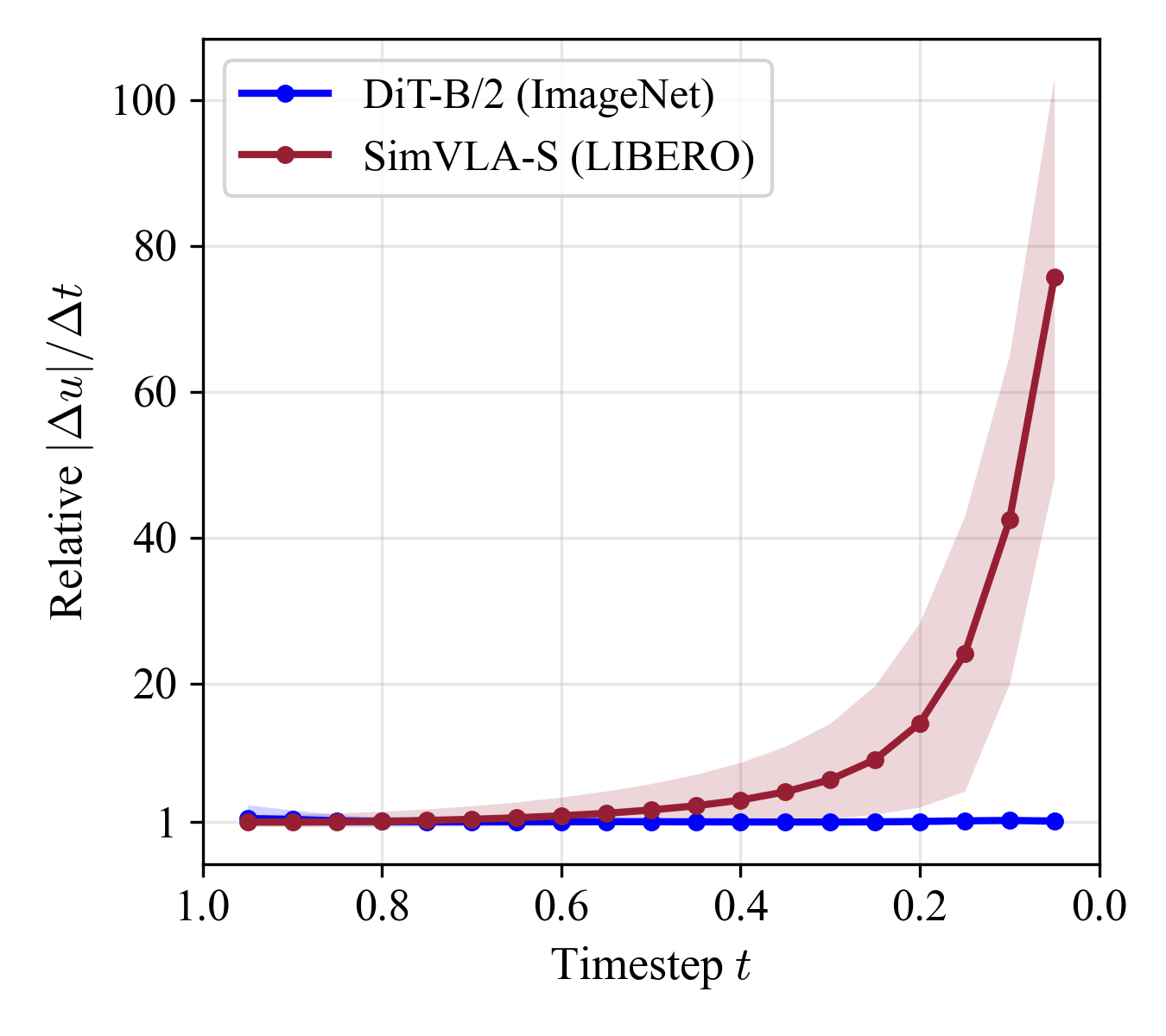}
        \caption{Velocity field comparison of two domains.}
        \label{fig:teaser:a}
    \end{subfigure}
    \hfill 
    \begin{subfigure}[b]{0.55\linewidth}
        \centering
        \includegraphics[width= 0.95\linewidth]{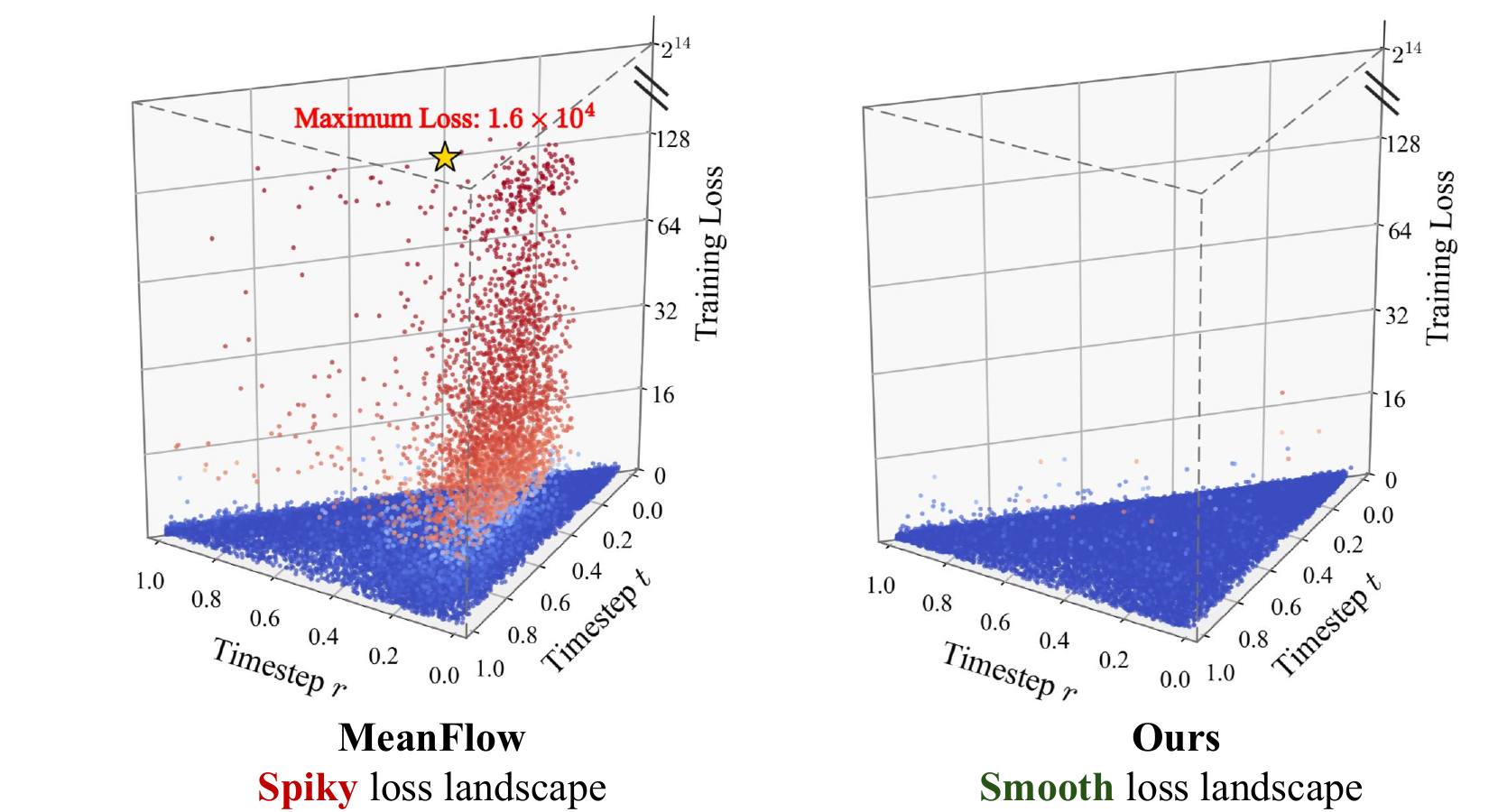} 
        \caption{Loss landscapes of MeanFlow and our K-MF.}
        \label{fig:teaser:b}
    \end{subfigure}
    \vspace{-2mm}
    \caption{\textbf{Illustrations of our key observations.} In (a), we calculate the change in average velocity $\Delta \mathbf{u}$ over a time interval of $\Delta t = 0.05$ and compare the dynamics of image generation (DiT-B on ImageNet) and action generation (SimVLA-S paired with DiT-B on LIBERO).
    To highlight the trend, we use the relative value to the initial value of each domain. The shaded area is bounded by the standard 1.5-IQR fences. In (b), we compare the loss landscapes of standard MeanFlow and our proposed K-MF, both applied to SimVLA-S on LIBERO. \textit{More examples are shown in Figure~\ref{fig:3}}.}
    \vspace{-5mm}
    \label{fig:teaser}
\end{figure}

\textbf{Observations and underlying issues.} In principle, the ground-truth average velocity between $t$ and $r$ is defined as the time-averaged integral of instantaneous velocities.
To avoid the costly integration, MeanFlow reformulates the learning target to a form that \textbf{evaluates locally} at timestep $t$: average velocity $\mathbf{u}_{t,r}$ equals the combination of instantaneous velocity $\mathbf{v}_t$ and the time derivative term $\frac{d}{dt} \mathbf{u}_{t,r}$ (see Eq. \ref{eq:mf_indentity}).
As $\mathbf{v}_t$ remains constant for a given sample during training, the time derivative term primarily drives the optimization. 
To investigate the challenges of adapting MeanFlow to RFMs, we compare the dynamics of ``local acceleration" ($\Delta \mathbf{u}/\Delta t$) between image generation and action generation in RFMs across denoising timesteps. As shown in Figure \ref{fig:teaser:a}, the local acceleration of image generation is stable throughout denoising with a narrow spread. In contrast, RFMs exhibit distinctive dynamics, characterized by two key observations with corresponding issues:
\begin{itemize}[itemsep=2pt,topsep=0pt,parsep=0pt,leftmargin=20pt]
    \item[(1)] The local acceleration surges sharply in the late denoising stage (i.e., $t < 0.3$), ultimately reaching $76\times$ its initial value at $t=1$. \textit{\textbf{Issue 1:}} The time derivative term tends to incur substantial estimation errors if the model fails to infer the global trend from local observations\footnote{Local acceleration reflects $\frac{d}{d t} \mathbf{u}_{t,r}$ ($t-r<\epsilon$), whose temporal stability decides local-global consistency.}. 
    \item[(2)] The spread of local acceleration magnitude widens as denoising proceeds (indicated by the expanding red shaded area). Our further analysis in Figure \ref{fig:3} reveals that the extent of this spread is strongly correlated with the task diversity of the RFM training data.
    \textit{\textbf{Issue 2:}} Minor estimation errors in the early denoising stage tend to be amplified as this spread widens. 
\end{itemize}
As a result, a direct adaptation of MeanFlow to RFMs yields an unstable, spiky loss landscape (Figure~\ref{fig:teaser:b}). 
Notably, this instability differs from that reported in prior work~\citep{geng2026mean,zhang2026alphaflow, sheng2026mp1}: \textbf{loss outliers in RFMs are strongly associated with sharp increases in local acceleration magnitude and its spread}, predominantly located at $r<0.3$. 
This highlights the necessity of correctly estimating\footnote{$\frac{d}{dt} \mathbf{u}_{t,r}$ is not a pre-defined ground truth, but is generated online through the policy $\pi_\theta$ being trained.} the time derivative term $\frac{d}{dt} \mathbf{u}_{t,r}$ in RFMs.

\textbf{Design insights and contributions.} Our method builds on a fundamental kinematic identity: for any intermediate point $c \in (r, t)$, the average velocity $\mathbf{u}_{t,r}$ is exactly equal to the convex combination of the average velocities over the two sub-intervals $\mathbf{u}_{t,c}$ and $\mathbf{u}_{c,r}$.
Leveraging this identity, we decouple the derivative $\frac{d}{dt} \mathbf{u}_{t,r}$ into two sub-interval terms, $\frac{d}{dt} \mathbf{u}_{t,c}$ and $\frac{d}{dc} \mathbf{u}_{c,r}$. This is primarily driven by two considerations: (1) for the first issue, the term $\frac{d}{dt} \mathbf{u}_{t,c}$ guarantees robust estimation in the early denoising stage, while $\frac{d}{dc} \mathbf{u}_{c,r}$ focuses exclusively on corrections in the late denoising stage; (2) for the second issue, a second estimation at timestep $c$ serves as an anchor to reduce the impact of the widening spread and thereby mitigate error amplification throughout the denoising process.
Given this decomposition, we further investigate how to determine the intermediate point $c$.  We empirically find that a fixed midpoint clearly outperforms uniform random sampling, despite the latter covering a broader range of interval partitions. This suggests that the placement of the estimation anchor, rather than partition diversity, is critical. 
We then introduce a learnable strategy that adaptively determines $c$ conditioned on $t$ and $r$, yielding the best performance among the considered strategies.
Building on these components, we propose a simple but effective one-step action generation policy tailored for RFMs, termed \textit{Kinematic MeanFlow} (K-MF). With K-MF, RFMs can generate actions in a single step, achieving comparable performance to the multi-step flow matching policy. This is validated by our systematic experiments across: (1) two training paradigms, including training from scratch and fine-tuning of pre-trained flow-matching RFMs; (2) different RFMs, including GR00T-N1.6 and SimVLA; (3) four datasets with varying training data scales, ranging from 658 to 87K trajectories; and (4) both Sim2Sim and Real2Sim evaluation settings.

\vspace{-1mm}

\section{Related Works}
\vspace{-0.5mm}
Diffusion and flow models~\citep{ho2020denoising, rombach2022high, lipmanflow} are initially designed for image generation. 
To improve inference efficiency, previous studies have extensively explored how to reduce the required sampling steps. 
One line of work focuses on deriving a few-step model from a pre-trained many-step baseline through score or flow distillation~\citep{salimans2022progressive, liu2023flow, yin2024one}. 
Another line of research relies on consistency models~\citep{song2023consistency, song2024improved}, which employ consistency constraints to ensure outputs along the same path share identical endpoints. Recently, MeanFlow models~\citep{geng2026mean, geng2026improved} have emerged as a compelling alternative, which enables one-step generation by predicting the average velocity between any two timesteps. Several studies introduce intermediate points into MeanFlow to enforce consistency across interval partitions~\citep{li2026trajectory,guo2025splitmeanflow} or facilitate curriculum learning~\citep{zhang2026alphaflow}. 
These studies primarily aim to improve training efficiency or mitigate gradient discrepancies of MeanFlow in image and audio generation tasks.
\textit{This has clear differences from our method in motivation, scope, and design}.

For specialist policies in robotics, Diffusion Policy~\citep{chi2025diffusion} pioneered the integration of diffusion algorithms into action generation.
Inspired by few-step algorithms in image tasks, FlowPolicy~\citep{zhang2025flowpolicy} and Maniflow~\citep{pmlr-v305-yan25a} set a consistency constraint for flow matching in imitation learning, while CP~\citep{ding2024consistency} adapts this concept to behavioral cloning and actor-critic reinforcement learning algorithms.
Similarly, several recent studies adapt MeanFlow in action generation for both imitation learning~\citep{sheng2026mp1} and reinforcement learning~\citep{zhan2026mean, wang2026one, zou2026one}, as we have discussed before.

Prevalent generalist RFMs~\citep{bjorck2025gr00t, black2025pi0, zheng2025x, yuan2026fast,ye2026world} widely adopt flow matching for action generation. However, in stark contrast to specialist policies, research on few/one-step action generation in RFMs remains underexplored. Prior works~\citep{chen2026mean, luan2026snapflow} typically apply methods developed for image generation directly to RFMs or restrict their evaluation to a limited range of tasks and models. \textit{Differently, we reveal the unique dynamics of RFM velocity fields, design K-MF accordingly, and comprehensively validate it across models, tasks, and training paradigms.}

\begin{figure}
    \centering
    \includegraphics[width=\linewidth]{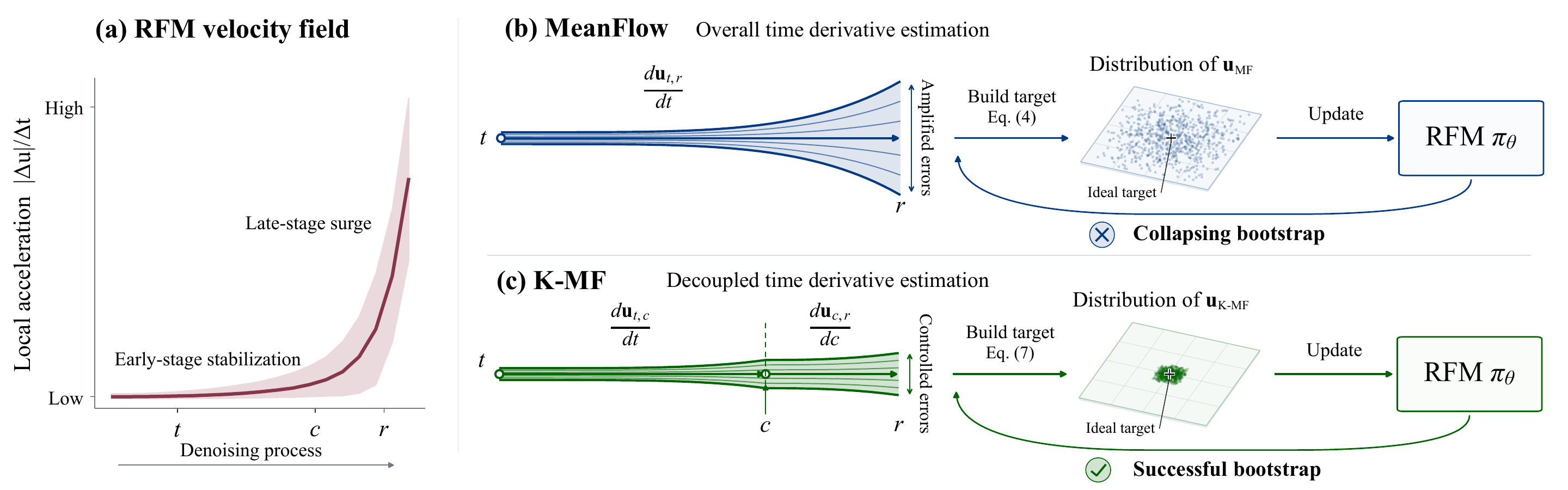}
    \caption{\textbf{Comparison of MeanFlow and our Kinematic MeanFlow (K-MF).} 
    In RFM velocity fields, the late-stage surge and widening spread shown in (a) lead to large errors in estimating the time derivative of average velocity over long intervals.
    These errors cause MeanFlow’s bootstrapping loop to collapse. In contrast, K-MF splits time-derivative estimation across two sub-intervals, mitigating these errors and stabilizing the bootstrapping process. 
    \textit{Note that the ideal target is a conceptual reference for better illustration. More detailed analysis is in Section \ref{sec:kmf}.}}
    \vspace{-3mm}
    \label{fig:method}
\end{figure}
\section{Method}
\label{sec:method}
\subsection{Preliminaries}
\textbf{Robotic foundation models with flow matching.} A typical RFM is trained to predict a sequence of future actions (i.e., an action chunk) by conditioning on current visual observation, language instruction, and proprioceptive state. Architecturally, RFMs mainly consist of two primary modules: (1) a backbone that encodes the visual observation and language instruction into a multimodal observation $\mathbf{o}_n$; (2) a transformer action head that generates the subsequent action chunk $\mathbf{a}_{n}$, conditioned on the proprioceptive state $\mathbf{s}_n$ and multimodal observation $\mathbf{o}_n$. 
For action generation, prevalent RFMs leverage flow matching~\citep{lipmanflow} to learn an instantaneous velocity field, modeling a continuous transformation from pure noise to valid actions. Specifically, 
let $\mathbf{a} \sim p_{\text{action}}$ denote a target action chunk and $\boldsymbol{\epsilon} \sim \mathcal{N}(\mathbf{0},\mathbf{I})$ denote Gaussian noise of the same dimensions as $\mathbf{a}$.
The flow path representing instantaneous velocity is constructed by: $\mathbf{z}_t = (1-t)\mathbf{a} + t\boldsymbol{\epsilon}$ with timestep $t\in[0,1]$. Correspondingly, the instantaneous velocity conditioned on the action chunk $\mathbf{a}$ at timestep $t$ is:
{\fontsize{9.3pt}{10.5pt}\selectfont
\begin{equation} \label{eq:instantaneous}
\mathbf{v}_t = d\mathbf{z}_t/dt = \boldsymbol{\epsilon} - \mathbf{a}.
\end{equation}
}%
\noindent The RFM $\pi_\theta$ is then trained to predict the instantaneous velocity by:
{\fontsize{9.3pt}{10.5pt}\selectfont
\begin{equation}
\mathcal{L}_{\text{FM}}(\theta) = \mathbb{E} \left[ \Vert  \pi_\theta(\mathbf{z}_t, t \mid \mathbf{o}, \mathbf{s}) - \mathbf{v}_t \Vert_2^2 \right].
\end{equation}
}%
During the inference stage, with the predicted velocity of $\pi_\theta$,
the RFM performs multiple sampling steps to gradually transform the noise $\boldsymbol{\epsilon}$ to the clean predicted action $\hat{\mathbf{a}}$.

\vspace{+1mm}

\textbf{Formulating MeanFlow in RFMs.} 
Unlike flow matching models that learn the instantaneous velocity, MeanFlow models~\citep{geng2026mean} establish a field that represents the average velocity between any two timesteps $t$ and $r$ ($0 \le r \le t \le 1$). The average velocity $\mathbf{u}$ is formulated as: $\mathbf{u}(\mathbf{z}_t, r, t) \triangleq \frac{1}{t - r} \int_{r}^{t} \mathbf{v}(\mathbf{z}_\tau, \tau) d\tau,$
where $\mathbf{v}(\mathbf{z}_\tau, \tau)$ denotes the marginal instantaneous velocity along the flow path $\mathbf{z}$ at timestep $\tau$. Alternatively, this equation can be rewritten as the following meanflow identity that describes the relation between instantaneous velocity  and average velocity:
{\fontsize{9.3pt}{10.5pt}\selectfont
\begin{equation} \label{eq:mf_indentity}
    \mathbf{u}(\mathbf{z}_t, r, t) = \mathbf{v}(\mathbf{z}_t, t) - (t - r) \frac{d}{dt} \mathbf{u}(\mathbf{z}_t, r, t).
\end{equation}
}%
During training, the RFM is optimized to predict $\mathbf{u}$ that satisfies Eq. \ref{eq:mf_indentity}. 
Following standard flow matching, the marginal instantaneous velocity $\mathbf{v}(z_t, t)$ is replaced by the conditional instantaneous velocity $\mathbf{v}_t$ for a given action chunk $\mathbf{a}$. 
The RFM $\pi_\theta$ is optimized with the learning objective:
{\fontsize{9.3pt}{10.5pt}\selectfont
\begin{equation}
\begin{aligned}
\mathcal{L}_{\text{MF}}(\theta) &= \mathbb{E} \left[ \left\| \pi_\theta(\mathbf{z}_t, r, t \mid \mathbf{o}, \mathbf{s}) - \mathbf{u}_{\text{MF}} \right\|_2^2 \right], \\
\text{where} \quad \mathbf{u}_{\text{MF}} &= \mathbf{v}_t - (t - r) \cdot \operatorname{sg}\left( \frac{d\pi_\theta(\mathbf{z}_t, r, t\mid \mathbf{o}, \mathbf{s})}{dt} \right), 
\end{aligned}
\label{eq:mf_loss}
\end{equation}
}%
\noindent and sg($\cdot$) denotes the stop-gradient operation. During the evaluation, MeanFlow uses the predicted average velocity to conduct denoising: $\mathbf{z}_r = \mathbf{z}_t - (t-r)\pi_\theta(\mathbf{z}_t, r,t\mid \mathbf{o}, \mathbf{s})$.
By simply setting $t=1$ and $r=0$, the one-step action generation can be achieved by: $\hat{\mathbf{a}} = \mathbf{z}_0 = \mathbf{z}_1 - \pi_\theta(\mathbf{z}_1,0,1\mid \mathbf{o}, \mathbf{s})$.

\subsection{Kinematic Meanflow}
\label{sec:kmf}
\textbf{Deeper analysis of time derivative term.} As stated in Section~\ref{sec:intro}, the primary bottleneck hindering MeanFlow in RFMs lies in the time derivative term. Specifically,  in Eq.~\ref{eq:mf_loss}, MeanFlow establishes a \textit{bootstrapping mechanism}:
at each training iteration, the derivative estimated by the current model is used to construct a stop-gradient regression target, and the updated model is then expected to provide more accurate derivatives for subsequent optimization.
Although this mechanism is empirically effective in image generation, we find that it breaks down in RFMs and identify two key contributing factors: (1) the late-stage surge in local acceleration makes derivative estimation prone to large errors; and (2) the widening spread tends to further amplify these errors (empirically demonstrated in Figure \ref{fig:expansion_ratio} in the Appendix). These errors corrupt the bootstrapping feedback loop and ultimately destabilize training, as illustrated in Figure \ref{fig:method}.
Consequently, MeanFlow struggles to achieve reliable one-step action generation in RFMs. We discuss this problem in detail in Section~\ref{sec:polit}.




\textbf{Decoupling time derivative via kinematic identity.} As revealed by the kinematic identity, for any intermediate point $c \in (r, t)$, the average velocity across the interval $[r, t]$ is identically equal to the convex combination of the average velocities across two partitioned sub-intervals, $[c, t]$ and $[r, c]$:
{\fontsize{9.3pt}{10.5pt}\selectfont
\begin{equation}
\begin{aligned}
    \mathbf{u}(\mathbf{z}_t, r, t) &= \frac{1}{t - r} \int_{r}^{t} \mathbf{v}(\mathbf{z}_\tau, \tau) \,\mathrm{d}\tau = \frac{1}{t - r} \left[ \int_{c}^{t} \mathbf{v}(\mathbf{z}_\tau, \tau) \,\mathrm{d}\tau + \int_{r}^{c} \mathbf{v}(\mathbf{z}_\tau, \tau) \,\mathrm{d}\tau \right] \\
    &= (1 - \lambda)\mathbf{u}(\mathbf{z}_t, c, t) + \lambda \mathbf{u}(\mathbf{z}_c, r, c), \quad \text{where } \lambda = \frac{c - r}{t - r}.
\end{aligned}
\label{eq:kinematic_identity}
\end{equation}
}%
Although interval partitioning is an established strategy in diffusion/flow models~\citep{salimans2022progressive, frans2025one, guo2025splitmeanflow,zhang2026alphaflow}, our method stands distinct in both underlying motivation and formulation from prior works.  
In contrast to prior works that construct a distillation target or improve training efficiency and stability in image generation, our method is motivated by the substantially different dynamics in RFM velocity fields and therefore decouples the time derivative term to address the resulting estimation difficulties. Specifically, we apply Eq.~\ref{eq:mf_indentity} separately to the two sub-intervals along the same flow path and obtain:
{\fontsize{9.3pt}{10.5pt}\selectfont
\begin{equation}
\begin{aligned}
    \mathbf{u}(\mathbf{z}_t, r, t) &= (1 - \lambda)\mathbf{u}(\mathbf{z}_t, c, t) + \lambda \mathbf{u}(\mathbf{z}_c, r, c)  \\
    &= (1 - \lambda)\left[\mathbf{v}(\mathbf{z}_t, t) - (t - c) \frac{d}{dt} \mathbf{u}(\mathbf{z}_t, c, t)\right] + \lambda \left[\mathbf{v}(\mathbf{z}_c, c) - (c - r) \frac{d}{dc} \mathbf{u}(\mathbf{z}_c, r, c)\right] \\
    &= \left[(1 - \lambda)\mathbf{v}(\mathbf{z}_t, t) + \lambda \mathbf{v}(\mathbf{z}_c, c)\right] - (t-r)\left[(1-\lambda)^2 \frac{d}{dt} \mathbf{u}(\mathbf{z}_t, c, t) + \lambda^2 \frac{d}{dc} \mathbf{u}(\mathbf{z}_c, r, c)\right].
\end{aligned}
\label{eq:kinematic_identity}
\end{equation}
}%
Given action chunk $\mathbf{a}$ during training,  both $\mathbf{z}_t$ and $\mathbf{z}_c$ in K-MF are constructed along the conditional flow path (see Section \ref{sec:discussion} in the Appendix for a detailed explanation). According to Eq. \ref{eq:instantaneous}, $\mathbf{v}(z_t, t)$ and $\mathbf{v}(\mathbf{z}_c, c)$ are replaced by $\mathbf{v}_t$ and $\mathbf{v}_c$, where $\mathbf{v}_t = \mathbf{v}_c = \boldsymbol{\epsilon}- \mathbf{a}$.
Formally, for the RFM policy $\pi_\theta$, the training objective of K-MF minimizes the mean-squared error against the decoupled target:
{{\fontsize{9.3pt}{10.5pt}\selectfont
\begin{equation}
\begin{gathered}
\mathcal{L}_{\text{K-MF}}(\theta) = \mathbb{E} \left[ \left\| \pi_\theta(\mathbf{z}_t, r, t \mid \mathbf{o}, \mathbf{s}) - \mathbf{u}_{\text{K-MF}} \right\|_2^2 \right], \\
\text{where} \ \  \mathbf{u}_{\text{K-MF}} = \mathbf{v}_t - (t - r) \left[ (1-\lambda)^2 \operatorname{sg}\left(\frac{d\pi_\theta(\mathbf{z}_t, c, t\mid \mathbf{o}, \mathbf{s})}{dt}\right) + \lambda^2 \operatorname{sg}\left(\frac{d\pi_\theta(\mathbf{z}_c, r, c\mid \mathbf{o}, \mathbf{s})}{dc}\right) \right].
\end{gathered}
\label{eq:kmf_loss}
\end{equation}
}

\textbf{Training objective with three strategies for determining $c$. } Based on the above analysis, the efficacy of K-MF largely relies on the second estimation of the time derivative at timestep $c$, which is dependent on the temporal ratio $\lambda$ via $c = r + \lambda (t - r)$. We then define three strategies.

\textit{Random sampling.} Under this strategy, the temporal ratio is sampled from a uniform distribution, $\lambda \sim \mathcal{U}(0, 1)$. By exposing the network to non-deterministic splits throughout training, this strategy serves as a baseline for assessing whether partition diversity benefits K-MF training.

\textit{Fixed at midpoint.} Under this strategy, we set the temporal ratio $\lambda = 0.5$ throughout training. In contrast to random sampling, this design completely eliminates variance introduced by interval splits and provides a balanced supervision signal across both sub-intervals.


\textit{Learnable strategy.} Under this strategy, the temporal ratio is generated by a learnable lightweight MLP (only 3 layers) throughout the training: $\lambda = G_\phi(t,r)$. 
The learning objective of $G_\phi(\cdot)$ is built upon the following three considerations: (1) an ideal temporal ratio $\lambda$ should not enlarge the training loss; instead, it should minimize the online bootstrapping error by balancing both sub-intervals; (2) in the early training stage, since the overall estimation $\frac{d}{dt} \mathbf{u}_{t,r}$ is inaccurate, $\lambda$ should be chosen to make the decoupled estimation $(1-\lambda)^2 \frac{d}{dt} \mathbf{u}_{t,c} + \lambda^2 \frac{d}{dc} \mathbf{u}_{c,r}$ distinct from the overall estimation; and (3) in the late training stage, as the bootstrapping feedback loop is well-established, the decoupled estimation should gradually approach the overall estimation to maintain the self-consistency. With these considerations, the learning objective of the learnable generator is:
{\fontsize{9.3pt}{10.5pt}\selectfont
\begin{equation}
\begin{aligned}
    \mathcal{L}_{\text{gen}}(\phi) &=  \mathbb{E} \left[ \omega \cdot \left\| \operatorname{sg}\left(\pi_\theta(\mathbf{z}_t, r, t \mid \mathbf{o}, \mathbf{s})\right) - \mathbf{u}_{\text{K-MF}}(\lambda) \right\|_2^2 \right], \ \  
    \text{where} \ \  \omega = 1 + \gamma e^{\left( -\frac{1}{\|\Delta \mathbf{u}\|_2 + \epsilon} \right)}.
\end{aligned}
\label{eq:generator_loss}
\end{equation}
}}%
$\Delta \mathbf{u}$ denotes the difference between the decoupled estimation and the overall estimation of each individual sample. With linearly increasing $\gamma$ from $-0.5$ to $0.5$ throughout training, the generated $\omega$ helps to achieve the second and third insights sequentially.  $\epsilon=10^{-6}$ to ensure numerical stability.

\textit{Overall training objective.} The overall training objective under each strategy is formulated as:
{\fontsize{9.3pt}{10.5pt}\selectfont
\begin{equation}
\mathcal{L} =
\begin{cases}
\mathcal{L}_{\text{K-MF}}(\theta),
& \text{random sampling / fixed at midpoint},\\[2pt]
\mathcal{L}_{\text{K-MF}}(\theta) + \alpha\mathcal{L}_{\text{gen}}(\phi),
& \text{learnable strategy},
\end{cases}
\end{equation}
}%
where $\alpha$ is the balancing factor of the two loss terms. 
During optimization, $\mathcal{L}_{\text{gen}}$ updates only the generator
parameters $\phi$, while the RFM parameters $\theta$ are treated as constants.
Notably, our formulation remains inherently straightforward. 
When adopting random sampling or the fixed-at-midpoint strategy, K-MF introduces no additional hyperparameters compared to MeanFlow. Under the learnable strategy, only two extra hyperparameters, $\alpha$ and $\gamma$, are introduced. 

\textbf{Implementation details of K-MF.} When implementing K-MF, we consider three aspects to balance training efficiency and performance. (1) \textit{Time derivative calculation:} following MeanFlow, we compute the time derivative via a Jacobian-vector product (JVP) between the network Jacobian $[\nabla_{\mathbf{z}}\pi_\theta, \partial_r \pi_\theta, \partial_t \pi_\theta]$ and the tangent vector $[\mathbf{v}_t, 0, 1]^\top$. By leveraging forward-mode automatic differentiation (dual numbers), each JVP is executed within a single forward pass without constructing a reverse computation graph. (2) \textit{Batched derivative calculation:} the two sub-interval derivative terms in Eq.~\ref{eq:kmf_loss} are mutually independent and operate under the stop-gradient operator. This decoupling allows us to concatenate the inputs of both sub-intervals along the batch dimension, evaluating both derivatives simultaneously in a single forward pass. (3) \textit{Adapting flow matching models for MeanFlow:} we introduce an additional embedding layer for the timestep $r$, and sum the embeddings of $t$ and $r$ to form the final time representation.
As a result, K-MF incurs no more than a 33\% increase in training time with acceptable GPU memory overhead relative to flow matching (see Table \ref{tab:training_overhead}).
The training and inference algorithms are shown in Algorithms~\ref{alg:kmf_train} and~\ref{alg:kmf_infer} in the Appendix, respectively.

\FloatBarrier
\begin{table*}[t]
\centering
\caption{\textbf{Pilot studies on MeanFlow and its variants on GR00T-N1.6.} In (a), we study the delicate training recipe for MeanFlow, including gradient clipping, progressive sampling for timesteps, and sufficient training iterations. In (b), we compare the performance between selected baselines and our K-MF (learnable strategy).
NFE denotes the number of function evaluations, SR is the success rate on the LIBERO-10 benchmark, and * denotes the delicate recipe. Best results are \textbf{bolded}.
}
\vspace{-2mm}
\begin{subtable}[t]{0.56\textwidth}
\centering
\caption{Delicate training recipe for MeanFlow.}
\label{tab:mf_training_recipe}
\vspace{-0.4em}
{\fontsize{7.2pt}{8pt}\selectfont
\setlength{\tabcolsep}{3.5pt}
\begin{NiceTabular}{l|c|c|cc|c}
\toprule
\textbf{Method} & \textbf{NFE} & \textbf{\# Iters} & \textbf{GradClip} & \textbf{Prog. Sampling} & \textbf{SR (\%)} \\
\midrule
\Block{2-1}{Flow Matching} & \Block{2-1}{4} & \Block{2-1}{20K} & & & 92.0 \\
& & & \checkmark & & 94.5 \\
\midrule
\Block{6-1}{MeanFlow} & \Block{2-1}{2} & \Block{2-1}{20K} & \checkmark & & \textit{Fail} \\
& & & \checkmark & \checkmark & 6.5 \\
\cmidrule{2-6}
& \Block{3-1}{2} & 40K & \checkmark & \checkmark & 52.5 \\
& & 60K & \checkmark & \checkmark & \textbf{87.0} \\
& & 80K & \checkmark & \checkmark & 84.5 \\
\cmidrule{2-6}
& 1 & 60K & \checkmark & \checkmark & \textit{Fail} \\
\bottomrule
\end{NiceTabular}
}
\end{subtable}
\hfill
\begin{subtable}[t]{0.41\textwidth}
\centering
\caption{Selected baselines \textit{vs.} K-MF}
\label{tab:baselines_vs_kmf}
\vspace{-0.4em}
\renewcommand{\arraystretch}{0.93}
{\fontsize{7.2pt}{8pt}\selectfont
\setlength{\tabcolsep}{4.2pt}
\begin{NiceTabular}{l|c|c|c}
\toprule
\textbf{Method} & \textbf{NFE} & \textbf{\# Iters} & \textbf{SR (\%)} \\
\midrule
Flow Matching & 4 & 20K & \textbf{94.5} \\
\midrule
\Block{2-1}{MeanFlow$^*$} & 2 & \Block{2-1}{60K} & 87.0 \\
& 1 & & \textit{Fail} \\
\midrule
\Block{2-1}{MVP} & 2 & \Block{2-1}{20K} & 82.0 \\
& 1 & & \textit{Fail} \\
\midrule
\Block{2-1}{$\alpha$-Flow} & 2 & \Block{2-1}{20K} & 94.0 \\
& 1 & & 84.5 \\
\midrule
\textbf{Ours} & 1 & 20K & \textbf{94.5} \\
\bottomrule
\end{NiceTabular}
}
\end{subtable}
\vspace{-0.2em}

\label{tab:combined_meanflow_analysis}
\end{table*}

\begin{table*}[t]
\centering
\caption{\textbf{Performance comparison across two training paradigms on the LIBERO benchmark.} We compare flow matching, MeanFlow (NFE=2), and our K-MF (learnable strategy, NFE=1) under both fine-tuning on GR00T-N1.6 and training from scratch on SimVLA-S (without pre-training on action data). * denotes the delicate recipe in Table~\ref{tab:mf_training_recipe}. Best results are \textbf{bolded}.}
\vspace{-0.5em}
\label{tab:model_comparison}
{\fontsize{7.2pt}{8pt}\selectfont
\setlength{\tabcolsep}{4.3pt}
\begin{NiceTabular}{l|c|c| l| c| c| cccc c}
\toprule
\textbf{Model} & \textbf{\# Params} & \textbf{Training Paradigm} & \textbf{Method} & \textbf{\# Iters} & \textbf{NFE} & \textbf{Spatial} & \textbf{Object} & \textbf{Goal} & \textbf{Long} & \textbf{Avg.} \\
\midrule
\Block{3-1}{GR00T-N1.6} & \Block{3-1}{3B} & \Block{3-1}{Fine-tuning} 
  & Flow Matching & 20K  & 4 & 97.5 & 98.5 & \textbf{98.5} & \textbf{94.5} & 97.3 \\
& & & MeanFlow*     & 60K  & 2 & 96.5             & 99.0             & 95.5          & 87.0          & 94.5 \\
& & & \textbf{Ours} & 20K & 1 & \textbf{99.5} & \textbf{99.5} & 98.0 & \textbf{94.5} & \textbf{97.9} \\
\midrule
\Block{3-1}{SimVLA-S} & \Block{3-1}{0.6B} & \Block{3-1}{Training from scratch} 
  & Flow Matching & 200K & 10 & \textbf{97.0} & 97.4 & 98.0 & 84.2 & 94.2 \\
& & & MeanFlow*     & 200K & 2  & 95.8          & 97.2 & 97.8 & 84.0 & 93.7 \\
& & & \textbf{Ours} & 200K & 1 & 96.4 & \textbf{97.8} & \textbf{98.4} & \textbf{87.8} & \textbf{95.1} \\
\bottomrule
\end{NiceTabular}
}
\vspace{-1em}
\end{table*}

\section{Experiments}
\subsection{Setups}
\textbf{Training and evaluation benchmarks.} Four training datasets have varying trajectory numbers: LIBERO~\citep{liu2023libero}, BridgeData V2~\citep{walke2023bridgedata}, Fractal~\citep{brohan2023rt}, and a COMPASS-generated~\citep{liu2025compass} point navigation dataset. They contain 2K, 53K, 87K, and 658 trajectories, respectively. Our evaluation comprises both Sim2Sim and Real2Sim settings. For Sim2Sim, we utilize LIBERO with its native simulator and PointNav powered by COMPASS. For Real2Sim, we evaluate on Fractal and BridgeData V2 via SimplerEnv~\citep{li24simpler}. Following established protocols, we run 200 and 500 rollouts for GR00T-N1.6 and SimVLA-S, respectively.

\textbf{Selected RFMs and baseline methods.} We select two representative RFMs: GR00T-N1.6~\citep{bjorck2025gr00t}  and SimVLA-S~\citep{luo2026simvla} for the evaluation with fine-tuning and training from scratch paradigms, respectively. We primarily compare against multi-step flow matching~\citep{lipmanflow} to show our method can achieve  better performance. We evaluate MeanFlow~\citep{geng2026mean} to justify the necessity of decoupling derivatives in RFMs. For a more comprehensive comparison, we select $\alpha$-Flow~\citep{zhang2026alphaflow} and MVP~\citep{zhan2026mean} as representative variants for interval partitioning and stabilized action generation, respectively.

\subsection{Pilot Experiments}
\label{sec:polit}
We investigate two critical questions in this part: (1) how to adapt MeanFlow effectively into generalist RFMs; and (2) how existing MeanFlow variants perform in RFM action generation, by performing pilot experiments on GR00T-N1.6 using the LIBERO-10 benchmark.

\begin{wrapfigure}{r}{0.38\columnwidth}
\centering
\vspace{-0.2em}
\includegraphics[width=0.90\linewidth]{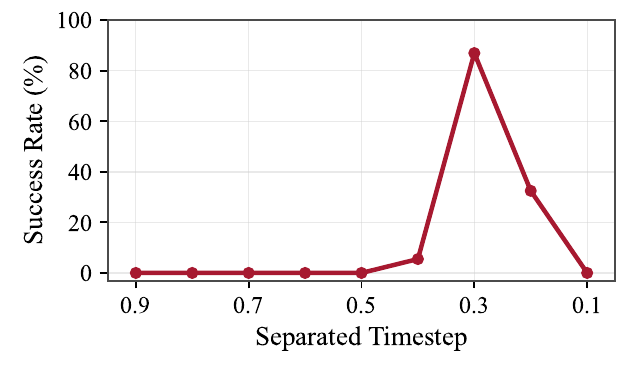}
\vspace{-1.2em}
\caption{\textbf{Sensitivity to the separated timestep} of two-step generation, under the best setting in Table~\ref{tab:mf_training_recipe}.}
\label{fig:mf_eval}
\vspace{-1.5em}
\end{wrapfigure}
\textbf{MeanFlow requires a delicate training recipe.} As illustrated in Table \ref{tab:mf_training_recipe}, besides the common gradient clipping, MeanFlow requires progressive sampling of timesteps (see Figure \ref{fig:progressive_sampling} in the Appendix) and sufficient training iterations. The performance saturates at around 60K iterations, then declines potentially due to overfitting. Even with the delicate training recipe, MeanFlow only reaches reasonable performance with two-step generation, still lagging behind standard flow matching. Meanwhile, its performance is sensitive to the separated timestep, as shown in Figure \ref{fig:mf_eval}.

\textbf{Performance of MeanFlow variants.} As shown in Table \ref{tab:baselines_vs_kmf}, $\alpha$-Flow and MVP do not require the delicate training recipe of MeanFlow, benefiting from interval-partitioning curriculum design and training stabilization via velocity constraints, respectively.
However, when pushed to one-step action generation, MVP fails to produce viable actions, and $\alpha$-Flow suffers noticeable performance degradation.
In contrast, our K-MF achieves one-step generation with comparable performance to multi-step flow matching. 

These results empirically support our hypothesis that the distinctive dynamics of the RFM velocity field pose challenges for modeling the average velocity, rendering existing methods insufficient to directly address one-step action generation in RFMs.

\subsection{Main Results}
We then examine the effectiveness of our method, as well as the generalization abilities across different training paradigms, RFMs, benchmarks, and embodiments.

\label{sec:exp_2}
\textbf{Different training paradigms and RFMs.} On the LIBERO benchmark, we perform fine-tuning on GR00T-N1.6 and training from scratch on SimVLA-S. As shown in Table \ref{tab:model_comparison}, K-MF consistently achieves competitive or superior performance to the multi-step flow matching and two-step MeanFlow using the delicate training recipe. This empirically validates that the efficacy of our K-MF is not limited to a specific training paradigm or RFM.

\textbf{Different benchmarks and embodiments.} We further fine-tune GR00T-N1.6 on large-scale tabletop manipulation benchmarks (Fractal and BridgeData V2) as well as the PointNav dataset. 
Table \ref{tab:exp_diff_task} shows that our K-MF consistently achieves higher success rates than the multi-step flow matching baseline across the majority of evaluated settings. These results demonstrate that K-MF generalizes across data scales, tasks, and robotic embodiments, extending from small-scale simulation data to large-scale real-world datasets and from tabletop manipulation to navigation.

\begin{table}[t]
\centering
\caption{\textbf{Performance comparison across benchmarks and embodiments.} We compare multi-step flow matching (FM) with our one-step K-MF across different fine-tuning datasets, evaluation settings, and robotic embodiments. Best results are \textbf{bolded}.}
\vspace{-2mm}
\renewcommand{\arraystretch}{0.90}
\label{tab:exp_diff_task}
{\fontsize{7.3pt}{8pt}\selectfont
\setlength{\tabcolsep}{7pt} 
\begin{NiceTabular}{l|l|l|l|c|cc}
\toprule
\Block{2-1}{\textbf{Task Type}} & \Block{2-1}{\textbf{Dataset}} & \Block{2-1}{\textbf{Setting}} & \Block{2-1}{\textbf{Embodiment}} & \Block{2-1}{\textbf{Task}} & \textbf{FM} & \textbf{Ours} \\
 & & & & & (NFE = 4) & (NFE = 1) \\
\midrule
\Block{10-1}{Tabletop\\Manipulation} 
 & \Block{5-1}{Fractal}  & \Block{5-1}{Real2Sim} & \Block{5-1}{GoogleX} 
 & Pick     & \textbf{91.0} & \textbf{91.0} \\
 & & & & Move     & 87.5          & \textbf{92.0} \\
 & & & & Open     & 35.5          & \textbf{50.5} \\
 & & & & Close    & \textbf{81.5} & 80.0          \\
\cmidrule{5-7}
 & & & & Avg.     & 73.9          & \textbf{78.4} \\
\cmidrule{2-7}
 & \Block{5-1}{BridgeData V2} & \Block{5-1}{Real2Sim} & \Block{5-1}{WidowX} 
 & Spoon    & 69.0          & \textbf{73.0} \\
 & & & & Carrot   & \textbf{64.0} & 60.0          \\
 & & & & Cube     & 7.5           & \textbf{13.0} \\
 & & & & Eggplant & 93.0          & \textbf{93.5} \\
\cmidrule{5-7}
 & & & & Avg.     & 58.4          & \textbf{59.9} \\
\midrule
\Block{3-1}{Point\\Navigation} 
 & \Block{3-1}{COMPASS\\Generated}  & \Block{3-1}{Sim2Sim}  & \Block{3-1}{Unitree G1} 
 & ID       &      \textbf{84.5}         & \textbf{84.5}     \\
 & & & & OOD      &     63.0          &   \textbf{63.5}            \\
\cmidrule{5-7}
 & & & & Avg.     &     73.8          &      \textbf{74.0}         \\
\bottomrule
\end{NiceTabular}
}
\vspace{-1mm}
\end{table}

\begin{figure*}[t]
\centering

\begin{minipage}[t]{0.61\textwidth}
\vspace{0pt}
\centering
\captionof{table}{\textbf{Ablation study on different strategies for determining $c$ on LIBERO}: random sampling (RS), fixed at midpoint (Mid), and learnable strategy (LS). Best and second-best results are \textbf{bolded} and \underline{underlined}.}
\vspace{-2mm}
\label{tab:selection_ablation}
{\fontsize{7.2pt}{8pt}\selectfont
\setlength{\tabcolsep}{4.2pt}
\begin{NiceTabular}{l|l|c|ccccc}
\toprule
\textbf{Model} & \textbf{Strategy} & \textbf{NFE}
& \textbf{Spatial} & \textbf{Object} & \textbf{Goal}
& \textbf{Long} & \textbf{Avg.} \\
\midrule
\Block{3-1}{GR00T-N1.6}
& RS  & 2 & \underline{98.0} & 99.0
& 96.5 & 87.5 & 95.3 \\
& Mid & 1 & \underline{98.0} & \textbf{99.5}
& \textbf{98.5} & \textbf{94.5} & \underline{97.6} \\
& LS  & 1 & \textbf{99.5} & \textbf{99.5}
& \underline{98.0} & \textbf{94.5} & \textbf{97.9} \\
\midrule
\Block{3-1}{SimVLA-S}
& RS  & 1 & 95.6 & 96.4
& \underline{97.6} & 85.6 & 93.8 \\
& Mid & 1 & \underline{96.2} & \underline{96.8}
& 97.2 & \underline{87.4} & \underline{94.4} \\
& LS  & 1 & \textbf{96.4} & \textbf{97.8}
& \textbf{98.4} & \textbf{87.8} & \textbf{95.1} \\
\bottomrule
\end{NiceTabular}
}
\end{minipage}\hfill
\begin{minipage}[t]{0.39\textwidth}
\vspace{0pt}
\centering
\includegraphics[width=0.73\linewidth]{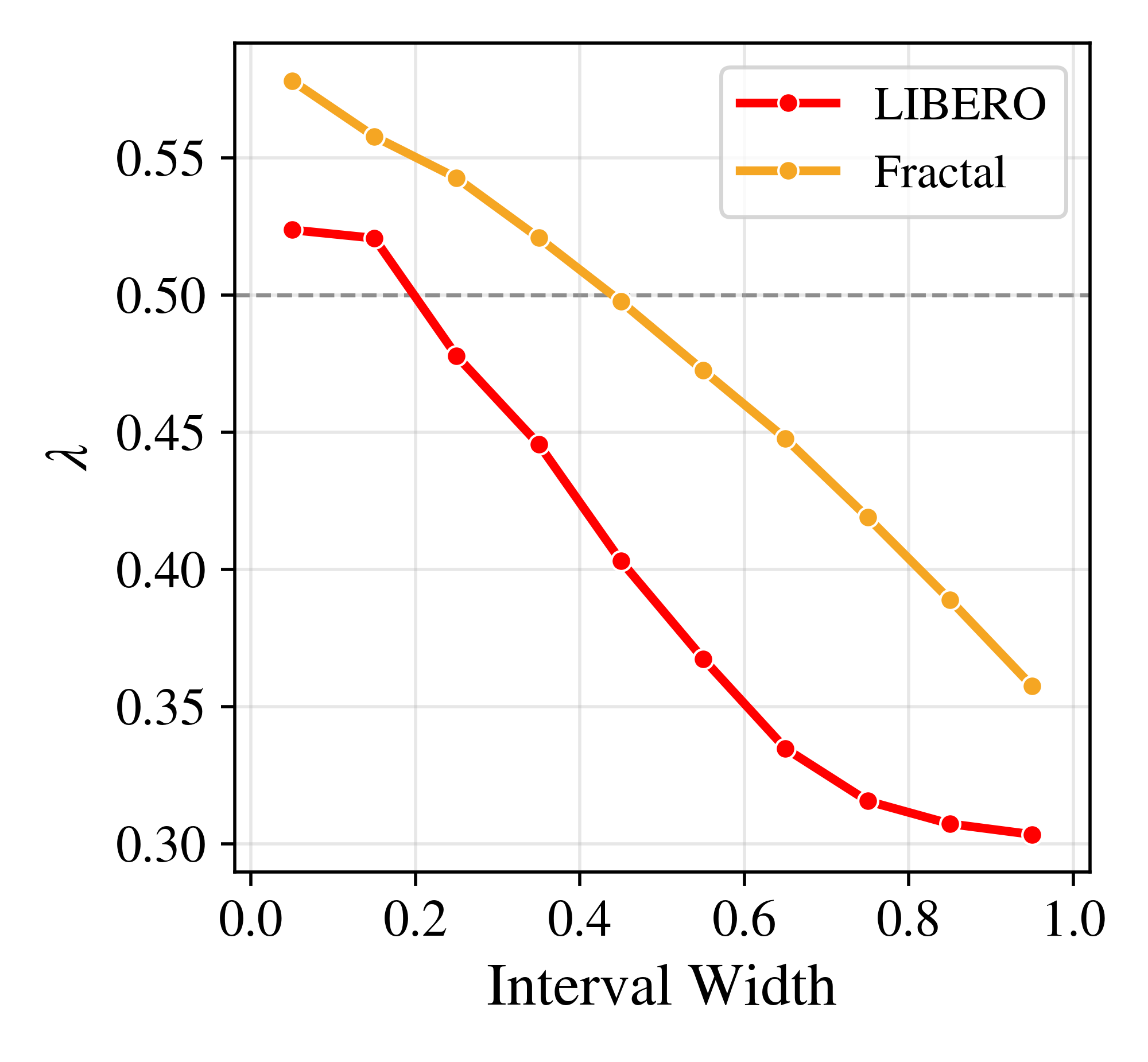}
\vspace{-3.8mm}
\caption{\textbf{Distribution of $\lambda$} for LS on GR00T-N1.6 across different datasets.}
\label{fig:lambda_LS}
\end{minipage}

\vspace{-5mm}
\end{figure*}

\vspace{-1mm}

\subsection{Ablation Studies}
\vspace{-1mm}
In this part, we first compare our three strategies for determining $c$ and then show the inference and training efficiencies of our method. 

\textbf{Different strategies for determining $c$.} We compare our three strategies in Table \ref{tab:selection_ablation}. The results indicate: (1) the \textit{learnable strategy} yields the best performance, and the \textit{fixed at midpoint} serves as a straightforward strategy for fast adaptation in the fine-tuning setting; (2) \textit{random sampling} yields clearly weaker performance than the other two strategies. Furthermore, as shown in Figure \ref{fig:lambda_LS}, we find that $\lambda$ in the learnable strategy shows strong correlations with the interval width $t-r$. These observations indicate that the efficacy of our K-MF primarily stems from improving time-derivative estimation rather than augmentation from different partitions.

\textbf{Inference efficiency.} 
As shown in Table~\ref{tab:performance_comparison}, we observe: (1) the action head accounts for 43\%--73\% of the end-to-end (E2E) latency of the flow matching baseline; (2) K-MF consistently reduces action-head latency by 67.5\%--74.4\% and E2E latency by 30.3\%--54.9\% across different hardware platforms and execution modes. Therefore, pairing generalist RFMs with K-MF largely facilitates their practical deployment on robotic platforms.

\textbf{Training overhead.} Table~\ref{tab:training_overhead} reports peak GPU memory usage and training time.
Compared to MeanFlow, K-MF incurs marginal additional overhead, increasing training time by 7.5\%--13.6\% and peak GPU memory by at most 2.5\%. In light of its inference efficiency, the training overhead of K-MF over flow matching remains acceptable: 33\% for GR00T-N1.6 and 27\% for SimVLA.

\textit{More ablation studies on hyperparameters and the learnable generator $G_\phi$ are in the Appendix.}

\begin{table*}[t]
\centering
\caption{\textbf{Efficiency analysis of K-MF.} In (a), we show the inference latency (ms) of flow matching and K-MF on GR00T-N1.6. AH and E2E denote action head and end-to-end latency, respectively. In (b), we show training overhead, in terms of peak GPU memory (GB) and training time (hours), under the same settings as in
Table~\ref{tab:model_comparison}, except keeping the same training iterations for fair comparison.}
\vspace{-2mm}
\label{tab:efficiency_overhead}
\begin{subtable}[t]{0.60\textwidth}
\vspace{0pt}
\centering
\caption{Inference latency.}
\vspace{-1.2mm}
\label{tab:performance_comparison}
\scriptsize
\setlength{\tabcolsep}{2.0pt}
\begin{tabular}{l|l|c|cc|cc}
\toprule
\multirow{2}{*}{Mode}
& \multirow{2}{*}{Platform}
& \multirow{2}{*}{Backbone}
& \multicolumn{2}{c|}{FM}
& \multicolumn{2}{c}{Ours} \\
& & & AH & E2E & AH & E2E \\
\midrule
\multirow{2}{*}{\textit{torch.eager}}
& L40
& 31.7 & 109.3 & 149.6
& 28.0$_{\downarrow 74.4\%}$
& 67.4$_{\downarrow 54.9\%}$ \\
& Jetson Orin
& 147.2 & 236.5 & 390.3
& 63.6$_{\downarrow 73.1\%}$
& 219.8$_{\downarrow 43.7\%}$ \\
\midrule
\multirow{2}{*}{\textit{torch.compile}}
& L40
& 19.4 & 19.1 & 44.2
& 6.2$_{\downarrow 67.5\%}$
& 30.8$_{\downarrow 30.3\%}$ \\
& Jetson Orin
& 70.4 & 82.8 & 162.9
& 23.3$_{\downarrow 71.9\%}$
& 102.8$_{\downarrow 36.9\%}$ \\
\bottomrule
\end{tabular}
\end{subtable}\hfill
\begin{subtable}[t]{0.38\textwidth}
\vspace{0pt}
\centering
\caption{Training overhead.}
\vspace{-1mm}
\label{tab:training_cost}
\label{tab:training_overhead}
\scriptsize
\setlength{\tabcolsep}{2.3pt}
\renewcommand{\arraystretch}{0.85}
\begin{tabular}{l|c|c|c|c}
\toprule
Model & Method & \# Iters & GPU Mem. & $T_{\mathrm{train}}$ \\
\midrule
\multirow{3}{*}{GR00T-N1.6}
& FM   & \multirow{3}{*}{20K}  & 35.6 & 9.4 \\
& MF &                      &  52.4 & 11.0 \\
& Ours &                      & 53.7 & 12.5 \\
\midrule
\multirow{3}{*}{SimVLA}
& FM   & \multirow{3}{*}{200K} & 22.8 & 58.6 \\
& MF   &  & 23.1 & 69.1 \\
& Ours &                      & 23.2 & 74.3 \\
\bottomrule
\end{tabular}
\end{subtable}
\vspace{-2mm}
\end{table*}

\subsection{Understanding the Dynamics in the Velocity Field of RFMs}
\begin{figure}[t]
    \centering
    \vspace{-1mm}
    \begin{subfigure}[b]{0.31\linewidth}
        \centering
        \includegraphics[width=0.95\linewidth]{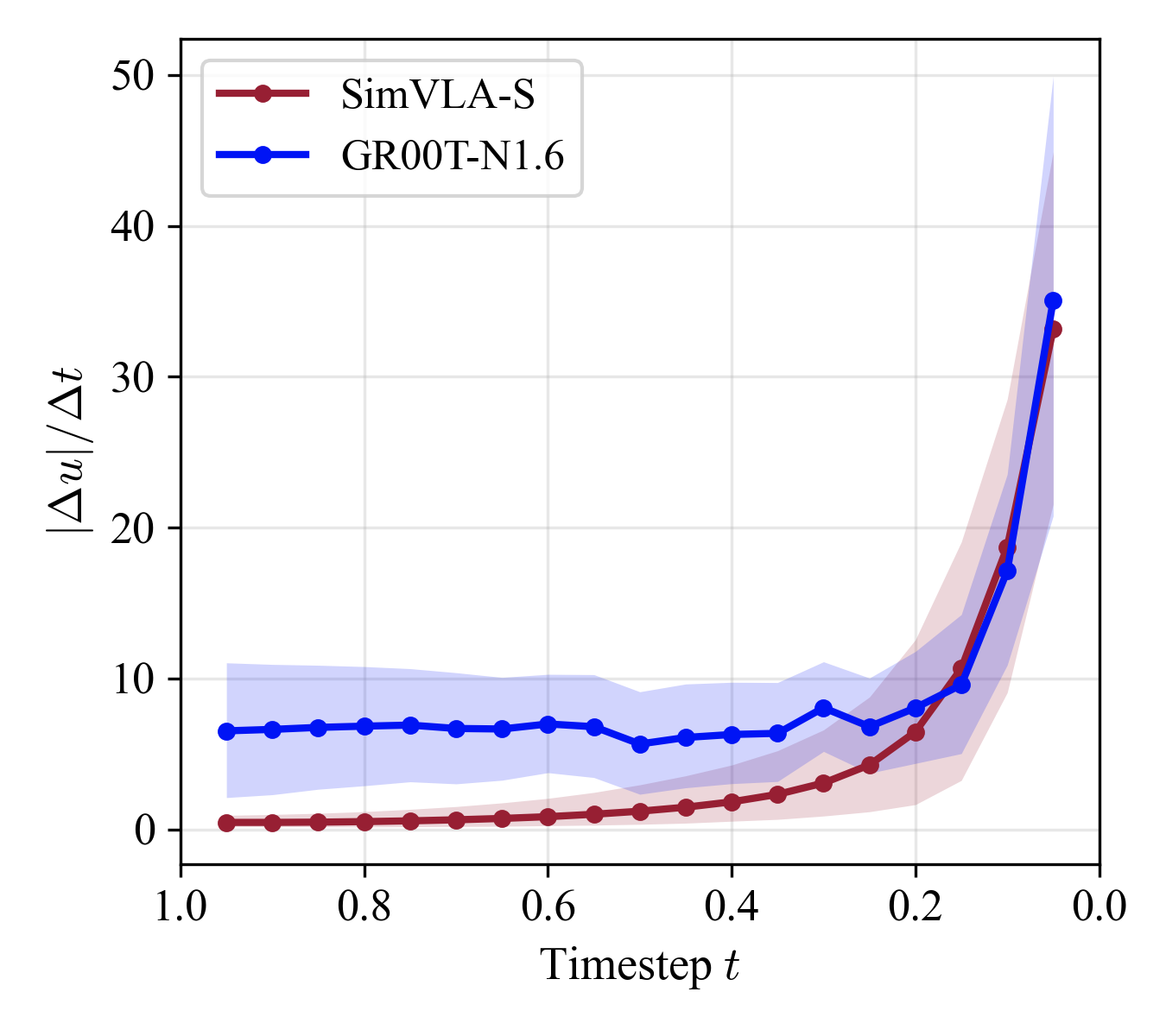}
        \vspace{-3mm}
        \caption{SimVLA-S \textit{vs.} GR00T-N1.6.}
        \label{fig:different_rfms}
    \end{subfigure}
    \hfill 
    \begin{subfigure}[b]{0.31\linewidth}
        \centering
        \includegraphics[width=0.95\linewidth]{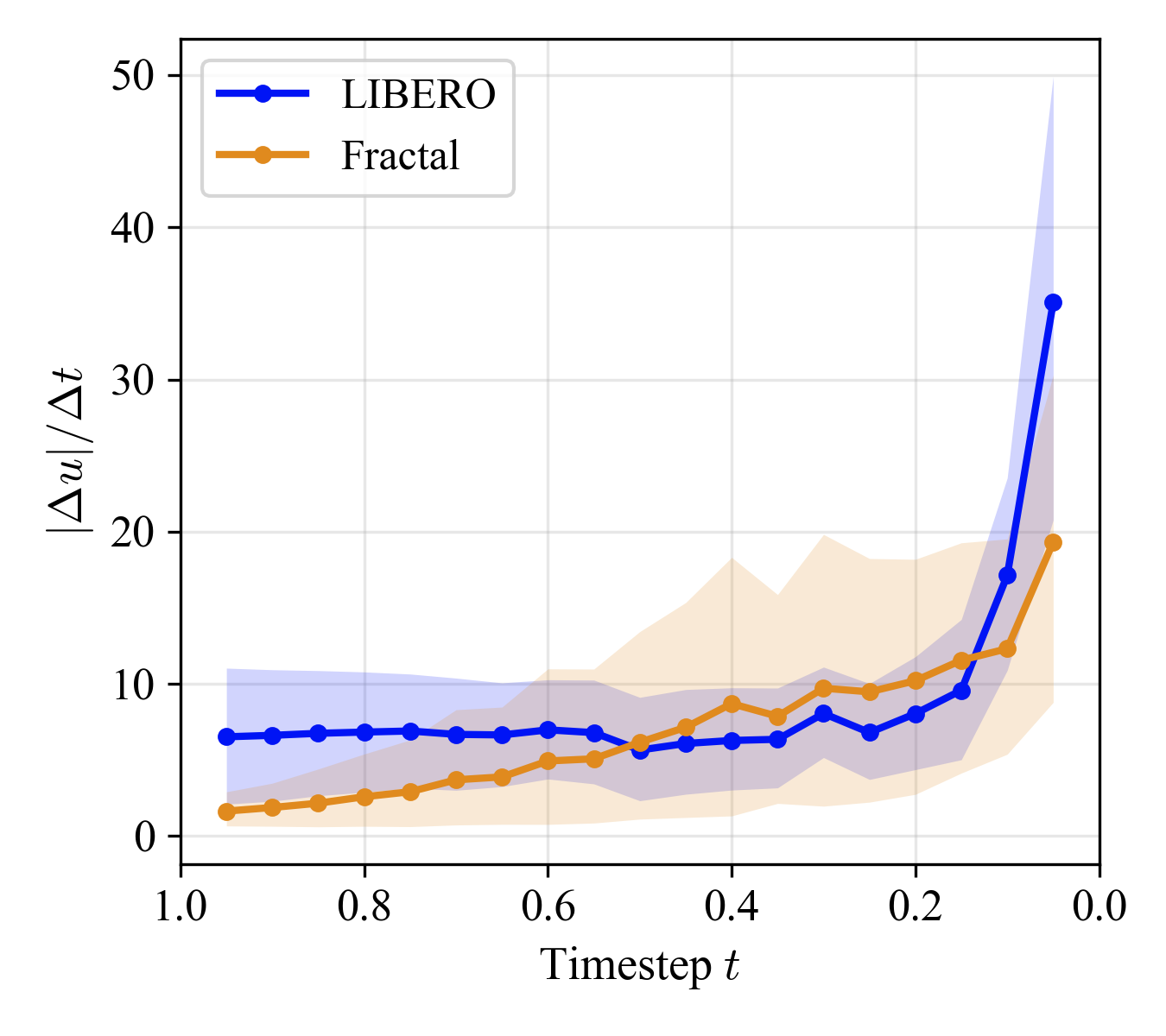}
        \vspace{-3mm}
        \caption{LIBERO \textit{vs.} Fractal.}
        \label{fig:different_dataset}
    \end{subfigure}
     \hfill 
    \begin{subfigure}[b]{0.33\linewidth}
        \centering
        \includegraphics[width=0.85\linewidth]{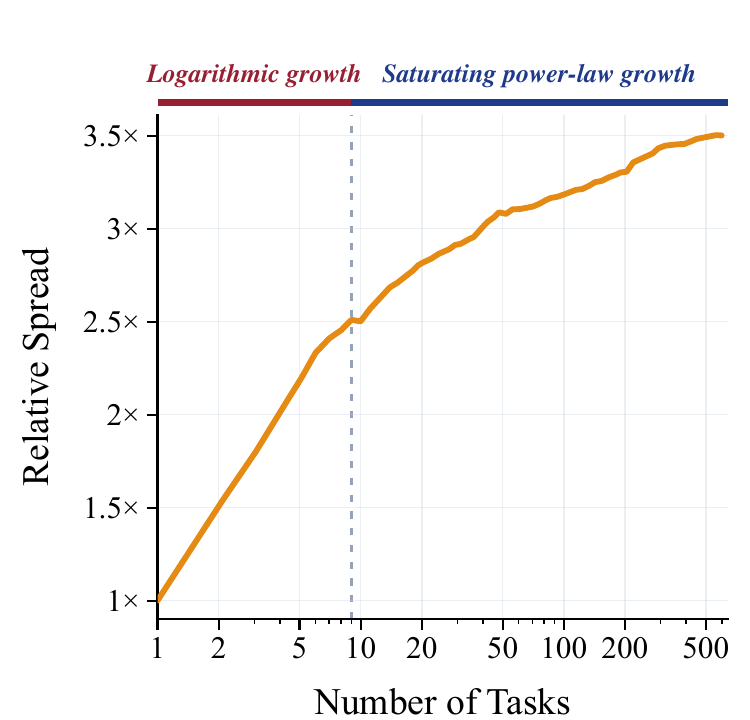}
        \vspace{-1mm}
        \caption{Relative spread \textit{vs.} \# tasks.}
        \label{fig:number_of_task}
    \end{subfigure}
    \vspace{-2mm}
    \caption{\textbf{More illustrations of the RFM velocity field.} (a) Comparison of dynamics between SimVLA-S and GR00T-N1.6 on LIBERO. (b) Dynamics of GR00T-N1.6 on LIBERO vs.~Fractal. (c) Relative spread of local acceleration magnitudes as the number of tasks increases.}
    \label{fig:3}
\end{figure}

Based on Figure \ref{fig:teaser:a}, we further investigate the RFM velocity field and have three observations: 
\begin{itemize}[itemsep=2pt,topsep=0pt,parsep=0pt,leftmargin=20pt]
    \item[(1)] \textbf{Different RFMs share a similar pattern on the same dataset.} As shown in Figure~\ref{fig:different_rfms}, GR00T-N1.6 and SimVLA-S exhibit a similar increasing trend in $|\Delta u|/\Delta t$ as denoising progresses, despite differences in their absolute values and the extent of spread.
    \item[(2)] \textbf{Dynamics depend critically on the training dataset.} As shown in Figure~\ref{fig:different_dataset}, the model trained on the larger and more complex Fractal dataset exhibits a wider spread in local acceleration magnitudes during the denoising process than that trained on LIBERO.
    \item[(3)] \textbf{Spread increases with task diversity.} As shown in Figure~\ref{fig:number_of_task}, the relative spread of local acceleration magnitudes within Fractal increases with task count, following an approximately logarithmic trend at low task counts and a saturating power-law trend at higher task counts.
    \end{itemize}
Based on these observations, we summarize two key takeaways. First, across architectures and datasets, the RFM velocity field exhibits a shared pattern, whereas the extent of its spread depends strongly on the training distribution. Second, higher task diversity is associated with a wider spread, suggesting that modeling average velocity becomes increasingly challenging as datasets grow larger and more diverse. \textit{More analysis is in Section \ref{sec:discussion} in the Appendix.}

\vspace{-1mm}

\section{Conclusion}
\vspace{-0.5mm}
In this paper, we present Kinematic MeanFlow (K-MF), a simple and effective framework for one-step action generation in generalist RFMs. To accommodate distinctive dynamics in the RFM velocity field, K-MF decomposes the overall time derivative into two sub-interval terms and further incorporates three strategies for determining intermediate points. Our experiments demonstrate that K-MF generally achieves superior performance to multi-step flow matching across diverse settings. 

\textbf{Limitations.} Compared to flow matching, K-MF indeed incurs additional training overhead, yet its overall training cost remains reasonable.
Furthermore, our evaluation does not cover whole-body control or dexterous manipulation, both of which are still challenging for current RFMs.


\newpage

\section*{Reproducibility Statement}
\label{sec:reproduce}
To facilitate reproducibility, we conduct all experiments using publicly available models and datasets. The main paper and appendix provide implementation details for our method and baselines, along with experimental settings, hyperparameters, and evaluation benchmarks. We will release our code and model checkpoints upon publication.

\section*{AI Use Statement}
\label{sec:llm}
As requested by the ICLR 2027 policy, we disclose the usage of AI tools. AI tools were used solely to assist with proofreading the manuscript.
We have manually reviewed all AI-assisted edits. The authors take full responsibility for the final content of this work.


\bibliography{iclr2027_conference}

@inproceedings{lipmanflow,
  title={Flow Matching for Generative Modeling},
  author={Lipman, Yaron and Chen, Ricky TQ and Ben-Hamu, Heli and Nickel, Maximilian and Le, Matthew},
  booktitle={ICLR},
  year={2023}
}

@inproceedings{geng2026mean,
  title={Mean flows for one-step generative modeling},
  author={Geng, Zhengyang and Deng, Mingyang and Bai, Xingjian and Kolter, J Zico and He, Kaiming},
  booktitle={NeurIPS},
  year={2025}
}

@article{bjorck2025gr00t,
  title={Gr00t n1: An open foundation model for generalist humanoid robots},
  author={NVIDIA and Bjorck, Johan and Casta{\~n}eda, Fernando and Cherniadev, Nikita and Da, Xingye and Ding, Runyu and Fan, Linxi and Fang, Yu and Fox, Dieter and Hu, Fengyuan and Huang, Spencer and others},
  journal={arXiv preprint arXiv:2503.14734},
  year={2025}
}

@inproceedings{brohan2023rt,
  title={RT-1: Robotics Transformer for Real-World Control at Scale},
  author={Brohan, Anthony and Brown, Noah and Carbajal, Justice and Chebotar, Yevgen and Dabis, Joseph and Finn, Chelsea and Gopalakrishnan, Keerthana and Hausman, Karol and Herzog, Alexander and Hsu, Jasmine and others},
  booktitle={RSS},
  year={2023},
}

@inproceedings{zitkovich2023rt,
  title={Rt-2: Vision-language-action models transfer web knowledge to robotic control},
  author={Zitkovich, Brianna and Yu, Tianhe and Xu, Sichun and Xu, Peng and Xiao, Ted and Xia, Fei and Wu, Jialin and Wohlhart, Paul and Welker, Stefan and Wahid, Ayzaan and others},
  booktitle={CoRL},
  year={2023}
}

@inproceedings{kim2025openvla,
  title={OpenVLA: An Open-Source Vision-Language-Action Model},
  author={Kim, Moo Jin and Pertsch, Karl and Karamcheti, Siddharth and Xiao, Ted and Balakrishna, Ashwin and Nair, Suraj and Rafailov, Rafael and Foster, Ethan P and Sanketi, Pannag R and Vuong, Quan and others},
  booktitle={CoRL},
  year={2024}
}

@inproceedings{black2025pi0,
  title={$\pi_0$: A vision-language-action flow model for general robot control},
  author={Black, Kevin and Brown, Noah and Driess, Danny and Esmail, Adnan and Equi, Michael Robert and Finn, Chelsea and Fusai, Niccolo and Groom, Lachy and others},
  booktitle={RSS},
  year={2025}
}

@inproceedings{zheng2025x,
  title={X-vla: Soft-prompted transformer as scalable cross-embodiment vision-language-action model},
  author={Zheng, Jinliang and Li, Jianxiong and Wang, Zhihao and Liu, Dongxiu and Kang, Xirui and Feng, Yuchun and Zheng, Yinan and Zou, Jiayin and Chen, Yilun and Zeng, Jia and others},
  booktitle={ICLR},
  year={2026}
}

@article{luo2026simvla,
  title={Simvla: A simple vla baseline for robotic manipulation},
  author={Luo, Yuankai and Chen, Woping and Liang, Tong and Wang, Baiqiao and Li, Zhenguo},
  journal={arXiv preprint arXiv:2602.18224},
  year={2026}
}

@inproceedings{ye2026world,
  title={World action models are zero-shot policies},
  author={Ye, Seonghyeon and Ge, Yunhao and Zheng, Kaiyuan and Gao, Shenyuan and Yu, Sihyun and Kurian, George and Indupuru, Suneel and Tan, You Liang and Zhu, Chuning and Xiang, Jiannan and others},
  booktitle={ICLRW},
  year={2026}
}

@article{yuan2026fast,
  title={Fast-wam: Do world action models need test-time future imagination?},
  author={Yuan, Tianyuan and Dong, Zibin and Liu, Yicheng and Zhao, Hang},
  journal={arXiv preprint arXiv:2603.16666},
  year={2026}
}

@inproceedings{song2023consistency,
  title={Consistency models},
  author={Song, Yang and Dhariwal, Prafulla and Chen, Mark and Sutskever, Ilya},
  booktitle={ICML},
  year={2023}
}

@inproceedings{song2024improved,
  title={Improved techniques for training consistency models},
  author={Song, Yang and Dhariwal, Prafulla},
  booktitle={ICLR},
  year={2024}
}

@inproceedings{liu2023flow,
title={Flow Straight and Fast: Learning to Generate and Transfer Data with Rectified Flow},
author={Xingchao Liu and Chengyue Gong and Qiang Liu},
booktitle={ICLR},
year={2023},
}

@inproceedings{yin2024one,
  title={One-step diffusion with distribution matching distillation},
  author={Yin, Tianwei and Gharbi, Micha{\"e}l and Zhang, Richard and Shechtman, Eli and Durand, Fredo and Freeman, William T and Park, Taesung},
  booktitle={CVPR},
  year={2024},
}

@inproceedings{salimans2022progressive,
  title={Progressive distillation for fast sampling of diffusion models},
  author={Salimans, Tim and Ho, Jonathan},
  booktitle={ICLR},
  year={2022}
}

@inproceedings{zhang2026alphaflow,
title={AlphaFlow: Understanding and Improving MeanFlow Models},
author={Huijie Zhang and Aliaksandr Siarohin and Willi Menapace and Michael Vasilkovsky and Sergey Tulyakov and Qing Qu and Ivan Skorokhodov},
booktitle={ICLR},
year={2026},
}

@inproceedings{ding2024consistency,
  title={Consistency models as a rich and efficient policy class for reinforcement learning},
  author={Ding, Zihan and Jin, Chi},
  booktitle={ICLR},
  year={2024}
}

@inproceedings{zhan2026mean,
  title={Mean flow policy with instantaneous velocity constraint for one-step action generation},
  author={Zhan, Guojian and Tao, Letian and Wang, Pengcheng and Wang, Yixiao and Li, Yiheng and Chen, Yuxin  and Li, Hongyang and Tomizuka, Masayoshi and Li, Shengbo Eben},
  booktitle={ICLR},
  year={2026}
}

@inproceedings{wang2026one,
  title={One-step generative policies with q-learning: A reformulation of meanflow},
  author={Wang, Zeyuan and Li, Da and Chen, Yulin and Shi, Ye and Bai, Liang and Yu, Tianyuan and Fu, Yanwei},
  booktitle={AAAI},
  year={2026}
}

@inproceedings{sheng2026mp1,
  title={Mp1: Meanflow tames policy learning in 1-step for robotic manipulation},
  author={Sheng, Juyi and Wang, Ziyi and Li, Peiming and Liu, Mengyuan},
  booktitle={AAAI},
  year={2026}
}

@inproceedings{pmlr-v305-yan25a,
  title = 	 {ManiFlow: A General Robot Manipulation Policy via Consistency Flow Training},
  author =       {Yan, Ge and Zhu, Jiyue and Deng, Yuquan and Yang, Shiqi and Qiu, Ri-Zhao and Cheng, Xuxin and Memmel, Marius and Krishna, Ranjay and Goyal, Ankit and Wang, Xiaolong and Fox, Dieter},
  booktitle = 	 {CoRL},
  year = 	 {2025},
}

@article{chen2026mean,
  title={Mean-flow based one-step vision-language-action},
  author={Chen, Yang and Ma, Xiaoguang and Zhao, Bin},
  journal={arXiv preprint arXiv:2603.01469},
  year={2026}
}

@article{luan2026snapflow,
  title={Snapflow: One-step action generation for flow-matching vlas via progressive self-distillation},
  author={Luan, Wuyang and Li, Junhui and Zhao, Weiguang and Zhang, Wenjian and Wu, Tieru and Ma, Rui},
  journal={arXiv preprint arXiv:2604.05656},
  year={2026}
}

@inproceedings{ho2020denoising,
  title={Denoising diffusion probabilistic models},
  author={Ho, Jonathan and Jain, Ajay and Abbeel, Pieter},
  booktitle={NeurIPS},
  year={2020}
}

@inproceedings{rombach2022high,
  title={High-resolution image synthesis with latent diffusion models},
  author={Rombach, Robin and Blattmann, Andreas and Lorenz, Dominik and Esser, Patrick and Ommer, Bj{\"o}rn},
  booktitle={CVPR},
  year={2022},
}

@inproceedings{chi2025diffusion,
  title={Diffusion policy: Visuomotor policy learning via action diffusion},
  author={Chi, Cheng and Feng, Siyuan and Du, Yilun and Xu, Zhenjia and Cousineau, Eric  and Burchfiel, Benjamin and Song, Shuran},
  booktitle={RSS},
  year={2023},
}

@inproceedings{zhang2025flowpolicy,
  title={Flowpolicy: Enabling fast and robust 3d flow-based policy via consistency flow matching for robot manipulation},
  author={Zhang, Qinglun and Liu, Zhen and Fan, Haoqiang and Liu, Guanghui and Zeng, Bing and Liu, Shuaicheng},
  booktitle={AAAI},
  year={2025}
}

@inproceedings{jang2022bc,
  title={Bc-z: Zero-shot task generalization with robotic imitation learning},
  author={Jang, Eric and Irpan, Alex and Khansari, Mohi and Kappler, Daniel and Ebert, Frederik and Lynch, Corey and Levine, Sergey and Finn, Chelsea},
  booktitle={CoRL},
  year={2021},
}

@article{zou2026one,
  title={One step is enough: Dispersive meanflow policy optimization},
  author={Zou, Guowei and Wang, Haitao and Wu, Hejun and Qian, Yukun and Wang, Yuhang and Li, Weibing},
  journal={arXiv preprint arXiv:2601.20701},
  year={2026}
}

@article{touvron2023llama,
  title={Llama: Open and efficient foundation language models},
  author={Touvron, Hugo and Lavril, Thibaut and Izacard, Gautier and Martinet, Xavier and Lachaux, Marie-Anne and Lacroix, Timoth{\'e}e and Rozi{\`e}re, Baptiste and Goyal, Naman and Hambro, Eric and Azhar, Faisal and others},
  journal={arXiv preprint arXiv:2302.13971},
  year={2023}
}

@article{touvron2023llama2,
  title={Llama 2: Open foundation and fine-tuned chat models},
  author={Touvron, Hugo and Martin, Louis and Stone, Kevin and Albert, Peter and Almahairi, Amjad and Babaei, Yasmine and Bashlykov, Nikolay and Batra, Soumya and Bhargava, Prajjwal and Bhosale, Shruti and others},
  journal={arXiv preprint arXiv:2307.09288},
  year={2023}
}

@article{beyer2024paligemma,
  title={Paligemma: A versatile 3b vlm for transfer},
  author={Beyer, Lucas and Steiner, Andreas and Pinto, Andr{\'e} Susano and Kolesnikov, Alexander and Wang, Xiao and Salz, Daniel and Neumann, Maxim and Alabdulmohsin, Ibrahim and Tschannen, Michael and Bugliarello, Emanuele and others},
  journal={arXiv preprint arXiv:2407.07726},
  year={2024}
}

@article{guo2025splitmeanflow,
  title={Splitmeanflow: Interval splitting consistency in few-step generative modeling},
  author={Guo, Yi and Wang, Wei and Yuan, Zhihang and Cao, Rong and Chen, Kuan and Chen, Zhengyang and Huo, Yuanyuan and Zhang, Yang and Wang, Yuping and Liu, Shouda and others},
  journal={arXiv preprint arXiv:2507.16884},
  year={2025}
}

@inproceedings{li2026trajectory,
  title={Trajectory consistency for one-step generation on euler mean flows},
  author={Li, Zhiqi and Sun, Yuchen and Chen, Duowen and He, Jinjin and Zhu, Bo},
  booktitle={ICML},
  year={2026}
}

@inproceedings{geng2026improved,
  title={Improved mean flows: On the challenges of fastforward generative models},
  author={Geng, Zhengyang and Lu, Yiyang and Wu, Zongze and Shechtman, Eli and Kolter, J Zico and He, Kaiming},
  booktitle={CVPR},
  year={2026}
}

@inproceedings{liu2023libero,
  title={Libero: Benchmarking knowledge transfer for lifelong robot learning},
  author={Liu, Bo and Zhu, Yifeng and Gao, Chongkai and Feng, Yihao and Liu, Qiang and Zhu, Yuke and Stone, Peter},
  booktitle={NeurIPS},
  year={2023}
}

@inproceedings{walke2023bridgedata,
    title={BridgeData V2: A Dataset for Robot Learning at Scale},
    author={Walke, Homer and Black, Kevin and Lee, Abraham and Kim, Moo Jin and Du, Max and Zheng, Chongyi and Zhao, Tony and Hansen-Estruch, Philippe and Vuong, Quan and He, Andre and Myers, Vivek and Fang, Kuan and Finn, Chelsea and Levine, Sergey},
    booktitle={CoRL},
    year={2023}
}

@article{liu2025compass,
      title={COMPASS: Cross-embodiment Mobility Policy via Residual RL and Skill Synthesis},
      author={Liu, Wei and Zhao, Huihua and Li, Chenran and Deng, Yuchen and Biswas, Joydeep and Pouya, Soha and Chang, Yan},
      journal={arXiv preprint arXiv:2502.16372},
      year={2025}
}

@inproceedings{frans2025one,
title={One Step Diffusion via Shortcut Models},
author={Kevin Frans and Danijar Hafner and Sergey Levine and Pieter Abbeel},
booktitle={ICLR},
year={2025},
}

@article{li24simpler,
         title={Evaluating Real-World Robot Manipulation Policies in Simulation},
         author={Xuanlin Li and Kyle Hsu and Jiayuan Gu and Karl Pertsch and Oier Mees and Homer Rich Walke and Chuyuan Fu and Ishikaa Lunawat and Isabel Sieh and Sean Kirmani and Sergey Levine and Jiajun Wu and Chelsea Finn and Hao Su and Quan Vuong and Ted Xiao},
         journal = {arXiv preprint arXiv:2405.05941},
         year={2024}
}
\bibliographystyle{iclr2027_conference}
\newpage
\appendix
\section*{Appendix}
\setcounter{table}{0}
\setcounter{figure}{0}
\renewcommand{\thetable}{\Alph{table}}
\renewcommand{\thefigure}{\Alph{figure}}

\begin{itemize}
    \item Section~\ref{sec:dataset}: Datasets used for training in our experiments.
    \item Section~\ref{sec:baseline}: Implementation details of baseline methods.
    \item Section~\ref{sec:setups}: Experimental setups of our training and evaluation.
    \item Section~\ref{sec:ablation}: More ablation studies regarding our K-MF.
    \item Section~\ref{sec:discussion}: Discussion on two important problems.
\end{itemize}

\section{Datasets}
\label{sec:dataset}
\textbf{LIBERO}~\citep{liu2023libero} is a simulation benchmark for robotic manipulation. Our paper uses four task suites: LIBERO-Spatial, LIBERO-Object, LIBERO-Goal, and LIBERO-Long, which emphasize spatial relationships, object variations, goal variations, and long-horizon executions, respectively. Each task suite contains 10 tasks with 50 trajectories per task, yielding 2K trajectories across 40 tasks.

\textbf{Fractal}~\citep{brohan2023rt} is a large-scale dataset of real-world robotic manipulation collected with Google robots. The publicly released version contains 87K trajectories covering 599 tasks with different language instructions. The dataset mainly covers diverse manipulation behaviors in kitchen environments, including picking and placing objects and opening and closing drawers, providing substantial variation in tasks, objects, and scenes.

\textbf{BridgeData V2}~\citep{ walke2023bridgedata} is a large-scale dataset of real-world manipulation collected with the WidowX robot. The version we used consists of 53K trajectories spanning 13 skills across 24 environments. Most trajectories are annotated with natural language instructions, and the dataset includes variations in objects, camera viewpoints, and workspace configurations.

\textbf{PointNav} dataset we used is released by NVIDIA. It contains synthetic navigation demonstrations generated using COMPASS~\citep{liu2025compass}. Specifically, it contains 658 trajectories of the Unitree G1 robot, each of which provides robot state (including robot speed, route information, and goal heading), egocentric RGB observations, and velocity commands as actions.

\section{Implementation Details of Baseline Methods}
\label{sec:baseline}
\subsection{Flow Matching}
Flow matching~\citep{lipmanflow} is originally implemented by our selected RFMs. We mainly follow the official settings, with minor adjustments to hyperparameters to reproduce the reported results. Detailed setups are shown in Section \ref{sec:setups}.
\subsection{MeanFlow}
\begin{figure}[t]
    \centering
    \includegraphics[width=0.8\linewidth]{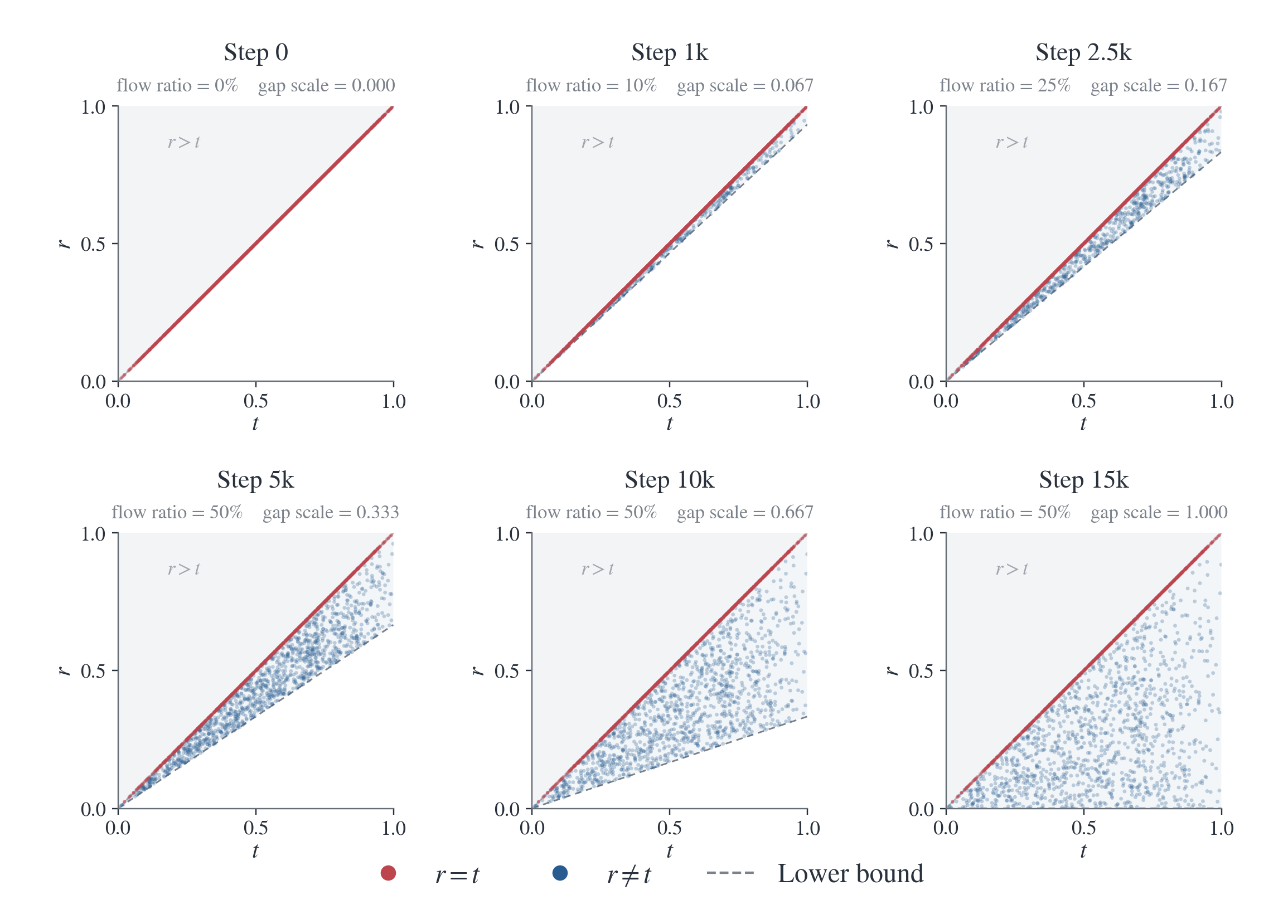}
    \vspace{-4mm}
    \caption{\textbf{Illustration of progressive sampling.} This strategy is governed by the flow ratio (which determines the proportion of $t\neq r$) and the gap scale (which defines the maximum interval for $t-r$). Initially, the training process reduces to traditional flow matching. As training progresses, the sampling of $t$ and $r$ gradually transitions to the standard formulation. \textit{Note that this strategy is exclusively applied to MeanFlow; our proposed K-MF does not require this technique.}}
    \label{fig:progressive_sampling}
\end{figure}

MeanFlow~\citep{geng2026mean} serves as the base method for our K-MF and shares its experimental setup, with the addition of progressive timestep sampling mentioned in Section~\ref{sec:polit} of the main paper. Specifically, progressive timestep sampling is crucial for training MeanFlow in RFMs. This strategy schedules two hyperparameters: the flow ratio and the gap scale. Specifically, the flow ratio increases linearly from 0 to 0.5, reducing the proportion of samples satisfying $t=r$ from 100\% to 50\%. The gap scale increases linearly from 0 to 1, progressively expanding the sampled temporal gaps from zero to their original magnitudes. By employing this scheduling, MeanFlow initially focuses on learning the local average velocity during the early stages of training before smoothly transitioning to the standard training objective. Figure~\ref{fig:progressive_sampling} illustrates this process.

\begin{table*}[t]
\centering
\renewcommand{\arraystretch}{1.15}
\caption{Hyperparameter settings of the flow matching baseline and our K-MF.}
\label{tab:exp_setups}
\fontsize{7.2pt}{8pt}\selectfont
\begin{tabular}{l l c c c c}
\toprule
\multirow{2}{*}{\textbf{Category}} & \multirow{2}{*}{\textbf{Hyperparameter}} & \textbf{SimVLA-S} & \multicolumn{3}{c}{\textbf{GR00T-N1.6}} \\
\cmidrule(lr){3-3} \cmidrule(lr){4-6}
\multicolumn{2}{l}{} & \textbf{LIBERO} & \textbf{LIBERO} & \textbf{Fractal/Bridge} & \textbf{PointNav} \\
\midrule
\multirow{12}{*}{\textbf{\textit{Shared}}} 
& Global batch size & 256             & 640             & 1024                    & 64 \\
& Learning rate     & 1.00e-04        & 1.00e-04        & 1.50e-04                & 1.00e-04 \\
& Optimizer         & AdamW           & AdamW           & AdamW                   & AdamW \\
& Betas             & (0.9, 0.95)     & (0.9, 0.95)     & (0.9, 0.95)             & (0.9, 0.95) \\
& Weight decay      & 0.0             & 1.00e-05        & 1.00e-05                & 1.00e-05 \\
& Training steps    & 200,000         & 20,000          & 30,000                  & 40,000 \\
& Warmup steps      & 1,000           & 2,000           & 1,500                   & 2,000 \\
& Grad clip         & 1.0             & 1.0             & 1.0                     & 1.0 \\
& Timestep Sampling & Beta(1.5,1.0)   & Beta(1.5,1.0)   & Beta(1.5,1.0)           & Beta(1.5,1.0) \\
& State drop rate   & 0.0             & 0.8             & 0.8                     & 0.0 \\
& Image resize      & 384$\times$384  & 224$\times$224  & 224 (min edge)          & 224 (min edge) \\
& Precision         & bf16            & bf16            & bf16                    & bf16 \\ 
\midrule
\multirow{2}{*}{\textbf{\textit{K-MF}}} 
& Flow ratio        & 0.5             & 0.5             & 0.5                     & 0.5 \\
& $\alpha$ & 0.1      & 0.1             & 0.1                     & 0.1 \\
& $\gamma$ & [-0.5, 0.5]      & [-0.5, 0.5]             & [-0.5,0.5]                     & [-0.5, 0.5] \\
\bottomrule
\end{tabular}
\end{table*}

\begin{table*}[t]
\centering
\caption{\textbf{Ablation studies on the LIBERO benchmark.}
We evaluate the hyperparameters $\alpha$ and $\gamma$ in the learnable strategy, the flow ratio inherited from MeanFlow (determining the proportion of $t \neq r$ per batch), and the depth of $G_\phi$.}
\begin{subtable}[t]{0.14\textwidth}
\vspace{0pt}
\centering
\caption{Effect of $\alpha$.}
{\fontsize{7.2pt}{8pt}\selectfont
\setlength{\tabcolsep}{4.2pt}
\begin{NiceTabular}{c|c}
\toprule
$\boldsymbol{\alpha}$ & \textbf{SR (\%)} \\
\midrule
0    & 97.6 \\
0.1  & 97.9 \\
0.5    & 97.6 \\
1.0    &  97.3 \\
2.0    &  96.3\\
\bottomrule
\end{NiceTabular}
\label{tab:ablation_alpha}
}
\end{subtable}\hfill
\begin{subtable}[t]{0.17\textwidth}
\vspace{0pt}
\centering
\caption{Effect of $\gamma$.}
{\fontsize{7.2pt}{8pt}\selectfont
\setlength{\tabcolsep}{4.2pt}
\begin{NiceTabular}{c|c}
\toprule
$\max{|\gamma|}$ & \textbf{SR (\%)} \\
\midrule
0            & 97.6 \\
0.25 & 97.8 \\
0.5   &  97.9\\
0.75   & 97.3 \\
1.0       & 96.1\\
\bottomrule
\end{NiceTabular}
\label{tab:ablation_gamma}
}
\end{subtable}\hfill
\begin{subtable}[t]{0.25\textwidth}
\vspace{0pt}
\centering
\caption{Effect of flow ratio.}
{\fontsize{7.2pt}{8pt}\selectfont
\setlength{\tabcolsep}{4.2pt}
\begin{NiceTabular}{c|c}
\toprule
Flow ratio & \textbf{SR (\%)} \\
\midrule
0.25          & 97.4  \\
0.5 & 97.9 \\
0.75   & 97.8 \\
1.00   & 97.5 \\
\bottomrule
\end{NiceTabular}
\label{tab:ablation_flow_ratio}
}
\end{subtable}\hfill
\begin{subtable}[t]{0.25\textwidth}
\vspace{0pt}
\centering
\caption{\# Layers of $G_\phi$.}
{\fontsize{7.2pt}{8pt}\selectfont
\setlength{\tabcolsep}{4.2pt}
\begin{NiceTabular}{c|c}
\toprule
\# Layers & \textbf{SR (\%)} \\
\midrule
2            & 97.3  \\
3          &  97.9  \\
4 &  97.8\\ 
\bottomrule
\end{NiceTabular}
\label{tab:ablation_layers}
}
\end{subtable}

\end{table*}

\subsection{Other MeanFlow Variants}
As we mentioned in the main paper, we incorporate $\alpha$-Flow~\citep{zhang2026alphaflow} and MVP~\citep{zhan2026mean} as representative MeanFlow variants for interval partitioning and stabilization techniques for action generation, respectively.

\textbf{$\alpha$-Flow.} Originally developed for image generation, $\alpha$-Flow aims to bridge the gap between the learning of instantaneous and average velocities through curriculum learning based on interval partitioning. Using its image-generation settings as a reference, we adjust the scheduling hyperparameters for action generation. Under this configuration, the model learns instantaneous velocity during the first 8,000 iterations and average velocity after 12,000 iterations. During the intervening 4,000 iterations, the model learns a combination of the two through interval partitioning, enabling a gradual transition.

\textbf{MVP.} This was initially designed for reinforcement learning. Here, we adopt only its instantaneous velocity constraint to stabilize training for action generation. Specifically, we retain the same training recipe as our flow matching baseline and augment the original MeanFlow objective with an auxiliary instantaneous velocity loss. Following the original implementation, we set the relative weight of this auxiliary loss to 1, assigning equal weights to the two losses.

\section{Experimental Setups}
\label{sec:setups}
\textbf{Training infrastructure.} All training experiments are conducted on a single node equipped with 8$\times$H20 GPUs (96GB).

\textbf{Codebase.} We implement K-MF and all baseline methods on top of the official codebases of GR00T-N1.6~\citep{bjorck2025gr00t} and SimVLA~\citep{luo2026simvla}.

\textbf{Hyperparameter configuration.} We mainly follow the standard training recipe. Table \ref{tab:exp_setups} summarizes the hyperparameter configuration adopted in flow matching and our K-MF in our experiments.

\textbf{Evaluation setups.} Our experiments are validated on three evaluation platforms, including LIBERO~\citep{liu2023libero}, SimplerEnv~\citep{li24simpler}, and COMPASS~\citep{liu2025compass}. 
All evaluations follow the standard protocol of the codebase.

\textit{LIBERO.} We evaluate on four benchmark suites: LIBERO-Spatial, LIBERO-Object, LIBERO-Goal, and LIBERO-Long (each containing 10 tasks, totaling 40 tasks). Following the original setting, we evaluate each task suite over 200 rollouts on GR00T-N1.6 and 500 rollouts on SimVLA-S.

\textit{SimplerEnv.} Models trained on Fractal and BridgeData V2 are evaluated on simulated GoogleX and WidowX embodiments, respectively. Following standard protocols, we evaluate GoogleX across diverse manipulation skills (``pick object'', ``move near'', ``open drawer'', and ``close drawer'') and WidowX across different target objects (``spoon on towel'', ``carrot on plate'', ``stack cube'', and ``put eggplant in basket''). Each task is evaluated over 200 rollouts.

\textit{COMPASS.} We evaluate navigation performance across both in-distribution (ID) and out-of-distribution (OOD) settings. Specifically, the ID setting uses the same environment seen during data collection, whereas the OOD setting adopts a novel ``hospital'' scenario as the representative benchmark. Each environment is evaluated over 200 rollouts.

\vspace{-1mm}
\section{More Ablation studies}
\vspace{-1mm}
\label{sec:ablation}
\label{sec:addtional ablation study}
\textbf{Hyperparameters $\alpha$ and $\gamma$.} K-MF introduces only two method-specific hyperparameters. 
$\alpha$ is the balancing factor of the loss for the learnable generator $G_{\phi}$ relative to the whole network, while $\gamma$ controls the penalty strength for large discrepancies between the overall and decoupled estimates. As shown in Table~\ref{tab:ablation_alpha}, choosing a relatively small $\alpha$ leads to better performance, with $\alpha=0.1$ achieving the best result.
Table~\ref{tab:ablation_gamma} further demonstrates that constraining the evolution of $\gamma$ within a moderate range, such as $[-0.5, 0.5]$, yields better performance.

\textbf{Flow ratio.} Inherited from MeanFlow, the flow ratio specifies the fraction of samples with $t\neq r$ in each training batch. As shown in Table~\ref{tab:ablation_flow_ratio}, K-MF is relatively insensitive to this hyperparameter within the tested range, maintaining strong performance even at a relatively high ratio of 0.75.

\begin{figure*}[t]
\centering
\begin{minipage}[t]{0.55\textwidth}
\renewcommand{\arraystretch}{1.10}
\vspace{0pt}
\centering
\captionof{table}{\textbf{Performance gain from learnable strategy across different evaluation benchmarks.} We use fixed at midpoint as the baseline and observe the relative performance gain.
Best results are \textbf{bolded}.}
{\fontsize{7.2pt}{8pt}\selectfont
\setlength{\tabcolsep}{4pt}
\begin{NiceTabular}{l|c|c|cc}
\toprule
\Block{2-1}{\textbf{Evaluation}\\\textbf{benchmark}}
& \Block{2-1}{\textbf{\# Training}\\\textbf{trajectories}}
& \Block{2-1}{\textbf{FM}\\\textbf{(NFE=4)}}
& \multicolumn{2}{c}{\textbf{Ours (NFE=1)}} \\
\cmidrule(lr){4-5}
& & & \textbf{Mid} & \textbf{LS} \\
\midrule
LIBERO         & 2K  & 97.3 & 97.6 & \textbf{97.9} \\
Simpler-WidowX & 53K & 58.4 & 59.1 & \textbf{59.9} \\
Simpler-Google & 87K & 73.9 & 76.8 & \textbf{78.4} \\
\bottomrule
\end{NiceTabular}
\label{tab:mid_vs_ls}
}
\end{minipage}
\hfill
\begin{minipage}[t]{0.44\textwidth}
\vspace{0pt}
\centering
\includegraphics[width=\linewidth]{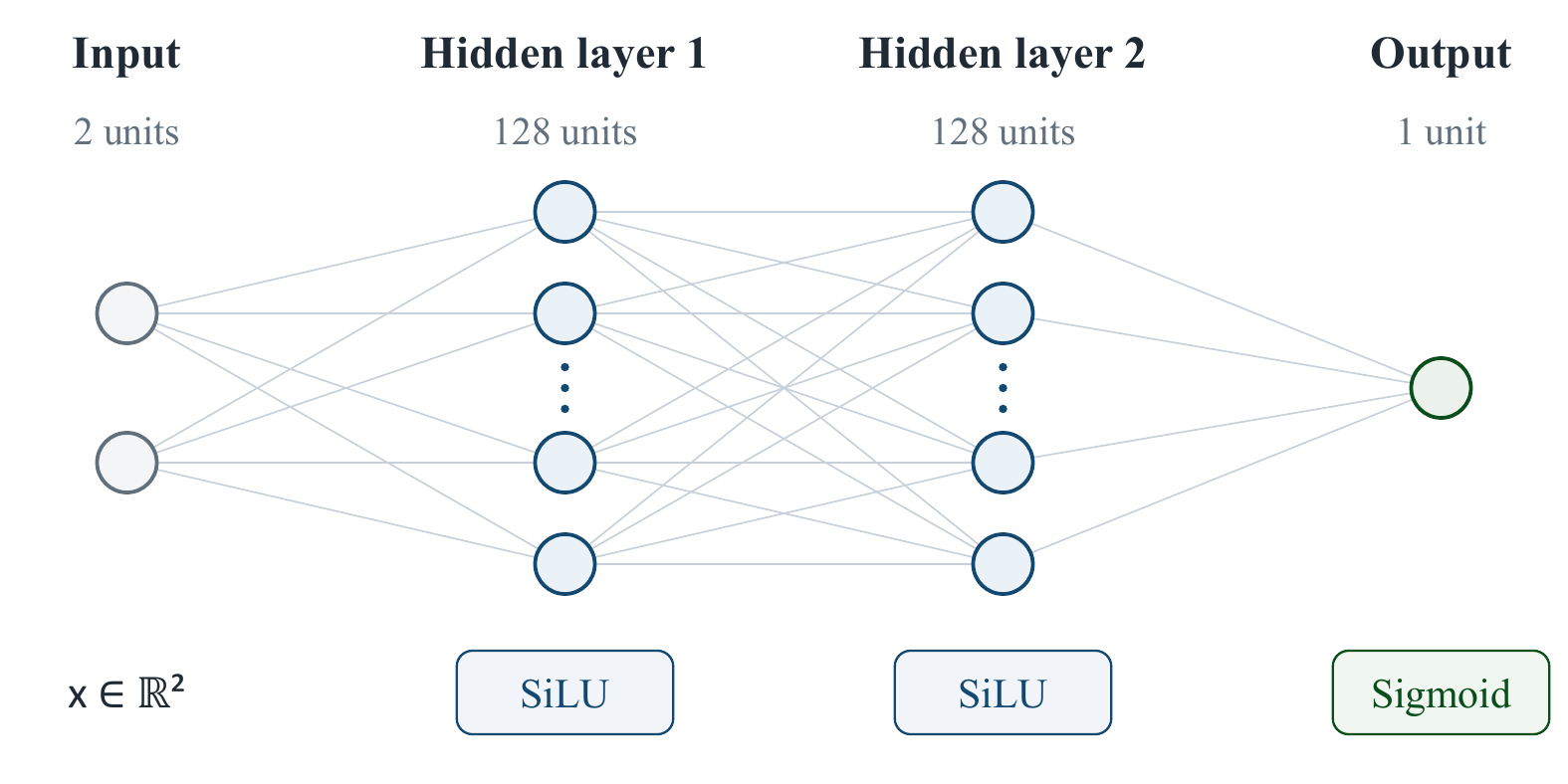}
\captionof{figure}{\textbf{The architecture of the learnable generator $G_\phi$.}}
\label{fig:learnable_generator}
\end{minipage}
\vspace{-3mm}
\end{figure*}

\begin{figure}[t]
    \centering
    \begin{subfigure}[b]{0.42\linewidth}
        \centering
\includegraphics[width=\linewidth]{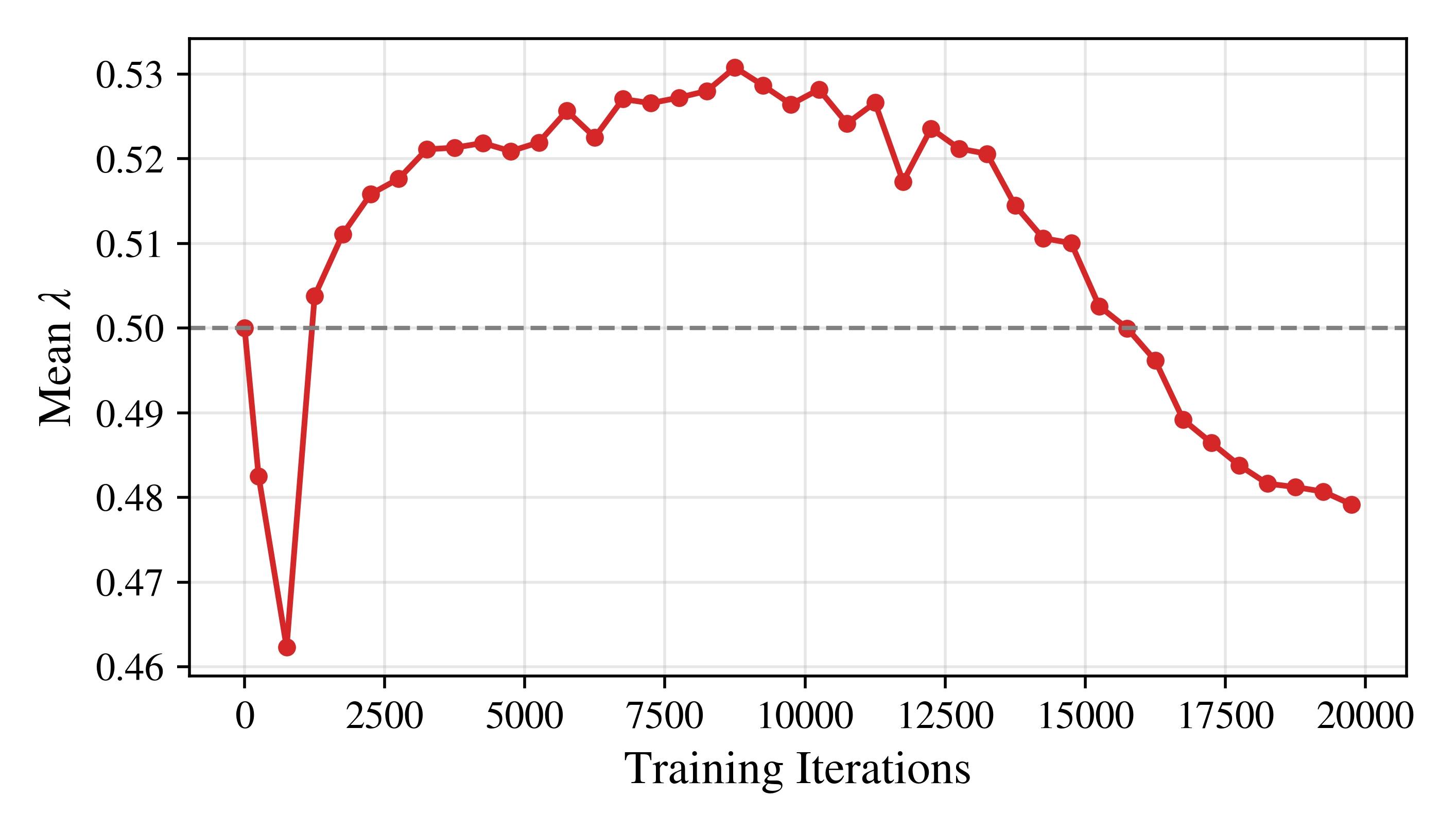}
\vspace{-7mm}
        \caption{Variation of mean $\lambda$ in training.}    \label{fig:libero_10_g_comparison}
    \end{subfigure}
    \begin{subfigure}[b]{0.50\linewidth}
        \centering \includegraphics[width=\linewidth]{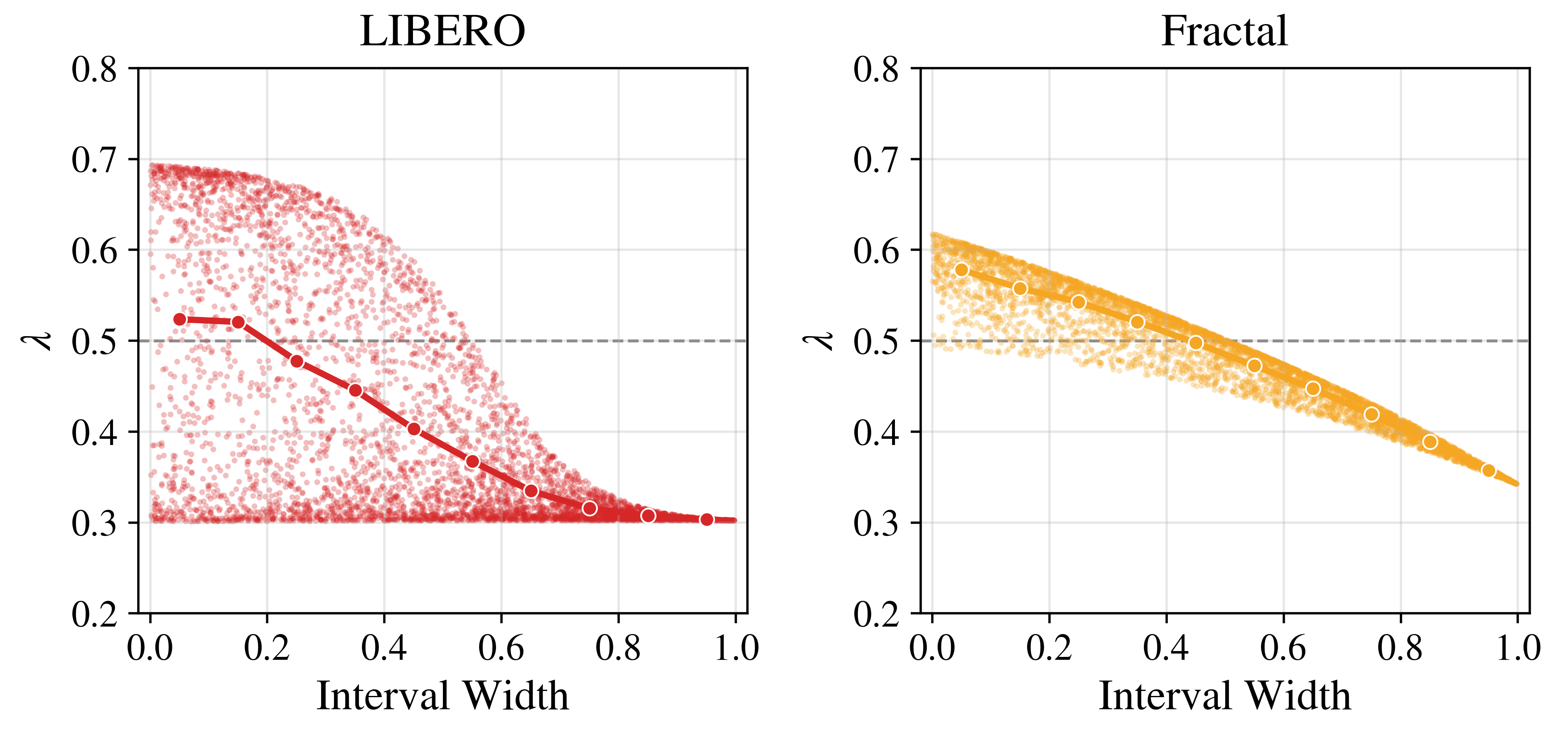}
        \caption{Distribution of $\lambda$ on different interval widths.}
        \label{fig:lambda_vs_width}
    \end{subfigure}
    \caption{\textbf{Ablation studies on learnable generator $G_\phi$ on GR00T-N1.6.} In (a), we illustrate the variation trend of the mean generated $\lambda$ throughout the training on LIBERO. In (b), we study the distribution of $\lambda$ generated by 
    the well-trained $G_\phi$ on LIBERO and Fractal datasets. \textit{As the four task suites in LIBERO share very similar properties, we use LIBERO-10 as the representative case.}}
\vspace{-2mm}
\end{figure}
\begin{figure}[t]
    \centering
\includegraphics[width=0.95\linewidth]{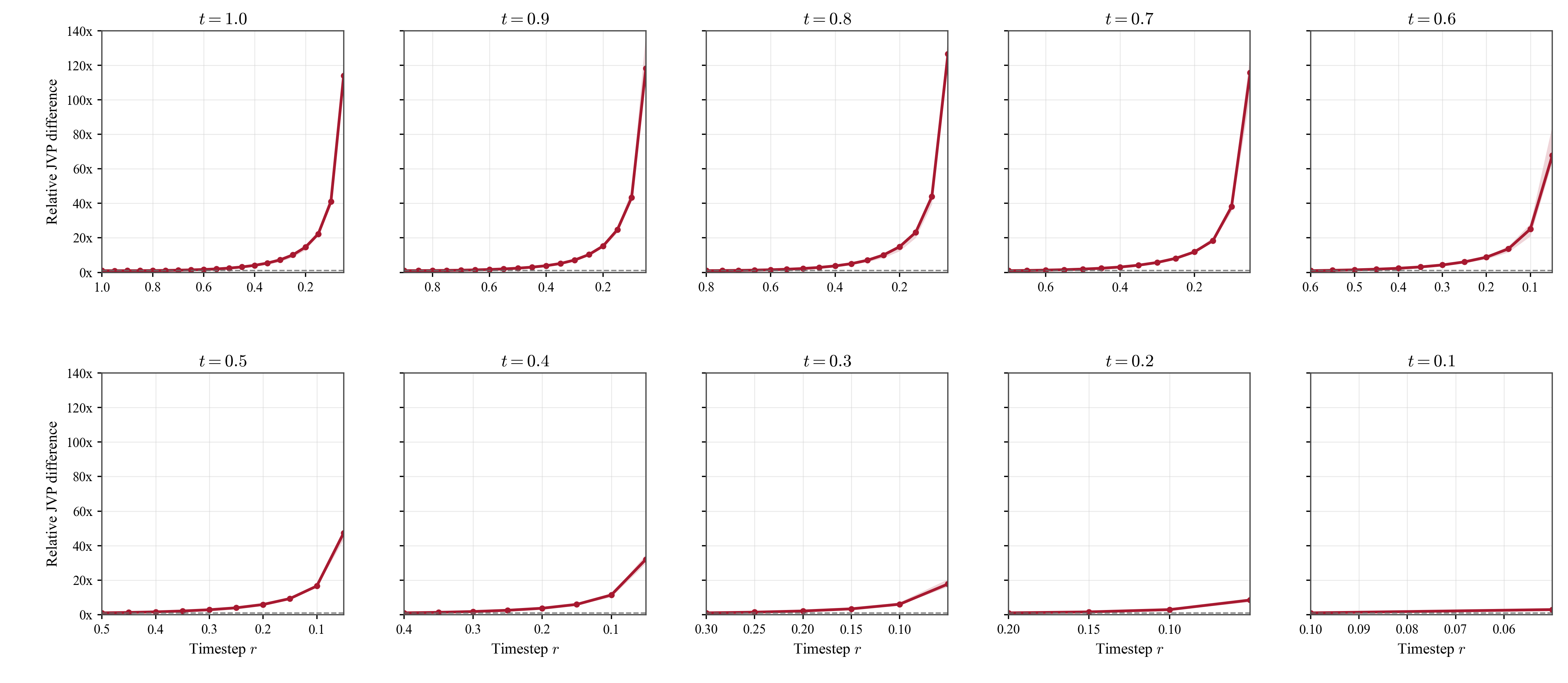}
\vspace{-2mm}
    \caption{\textbf{Relative error amplification across denoising timesteps.} On SimVLA-S, we introduce a 1\% random perturbation on $\mathbf{z}_t$, where $t\in\{1,0.9,\ldots,0.1\}$, and measure the relative amplification to the initial perturbation in $\frac{d\mathbf{u}_{t,r}}{dt}$ at $r\leq t$. The curve shows the mean value over 50 random trials.}
    \label{fig:expansion_ratio}
    \vspace{-2mm}
\end{figure}

\textbf{Learnable generator $G_\phi$.} As illustrated in Figure \ref{fig:learnable_generator}, $G_\phi$ is a three-layer MLP with layer dimensions $2 \rightarrow 128 \rightarrow 128 \rightarrow 1$. The first two layers are paired with SiLU, while the last layer is processed by Sigmoid, mapping the output into $(0,1)$\footnote{To prevent $G_\phi$ from degenerating over narrow time intervals, we impose boundary constraints on the sampling distribution on LIBERO.}. 
First, we ablate the number of layers in $G_\phi$, keeping the hidden dimension fixed at 128. As shown in Table~\ref{tab:ablation_layers}, the success rate increases from 97.3\% with two layers to 97.9\% with three layers, while adding a fourth layer yields no further improvement. We therefore adopt a three-layer MLP for $G_\phi$.
Second, we illustrate how the generated $\lambda$ evolves during training in Figure~\ref{fig:libero_10_g_comparison}. The mean generated $\lambda$ initially decreases and then rises during early training when $\gamma<0$. During later training when $\gamma>0$, it decreases and eventually converges to approximately 0.48.
Third, we visualize the distributions of $\lambda$ generated by the trained model across datasets in Figure~\ref{fig:lambda_vs_width}. We observe a consistent overall trend across datasets, with more concentrated $\lambda$ distributions on larger datasets. Lastly, we evaluate the additional performance gains from $G_\phi$ in Table~\ref{tab:mid_vs_ls}. We observe greater gains from the learnable generator on the two larger datasets, Fractal and BridgeData V2.

\textbf{Error amplification in late denoising stage.} As discussed in the main paper, errors in estimating the time derivative tend to be amplified as the denoising interval extends into later stages.  We perform an ablation study to validate this hypothesis. As illustrated in Figure \ref{fig:expansion_ratio}, a 1\% perturbation introduced at early denoising timesteps produces an error in $\frac{\mathrm{d}\mathbf{u}_{t,r}}{\mathrm{d}t}$ whose magnitude is approximately $120\times$ that of the initial perturbation by the end of denoising. This amplification decreases when the perturbation is introduced at later timesteps ($t<0.6$). 
Our K-MF decouples the overall time derivative into two separate terms during training, mitigating this error amplification based on three key facts: 
(1) in the early stages of denoising, error amplification remains mild; 
(2) in the late denoising phase (where $t$ is small), the propagation and amplification of noise are substantially suppressed; and 
(3) the two decomposed terms are computed independently along the conditional flow path, ensuring that errors accumulate additively rather than compounding multiplicatively.

\section{Discussion}
\label{sec:discussion}
\begin{figure}[t]
    \centering
    \begin{subfigure}[b]{0.88\textwidth}
        \centering
        \includegraphics[width=\linewidth]{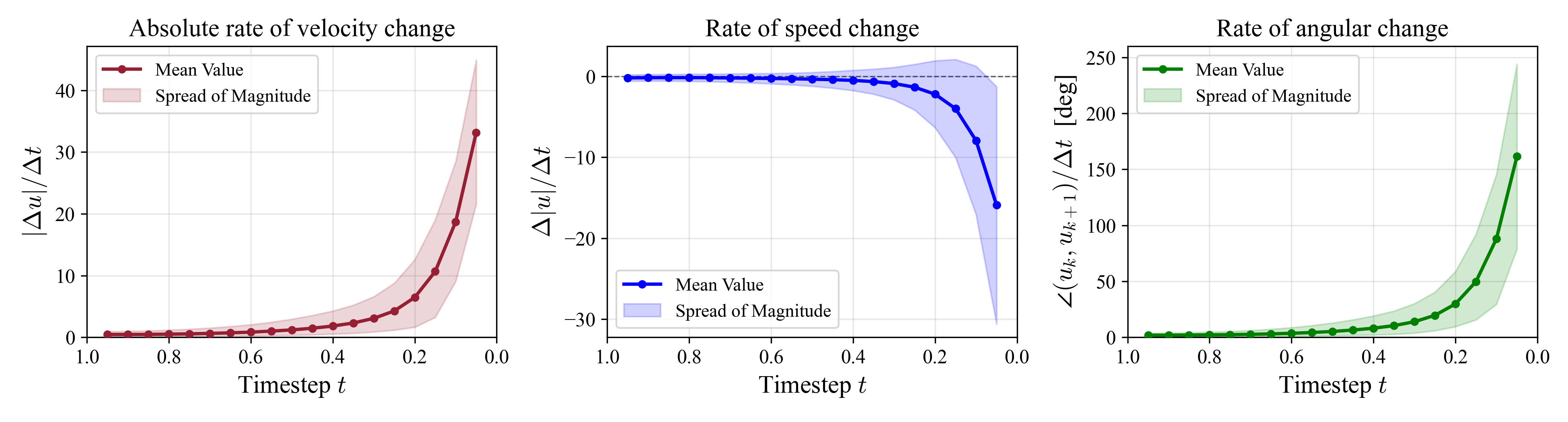}
        \caption{SimVLA-S paired with DiT-B on the LIBERO dataset.}
        \label{fig:mf_dynamics}
    \end{subfigure}
    \begin{subfigure}[b]{0.88\textwidth}
        \centering
        \includegraphics[width=1.0\linewidth]{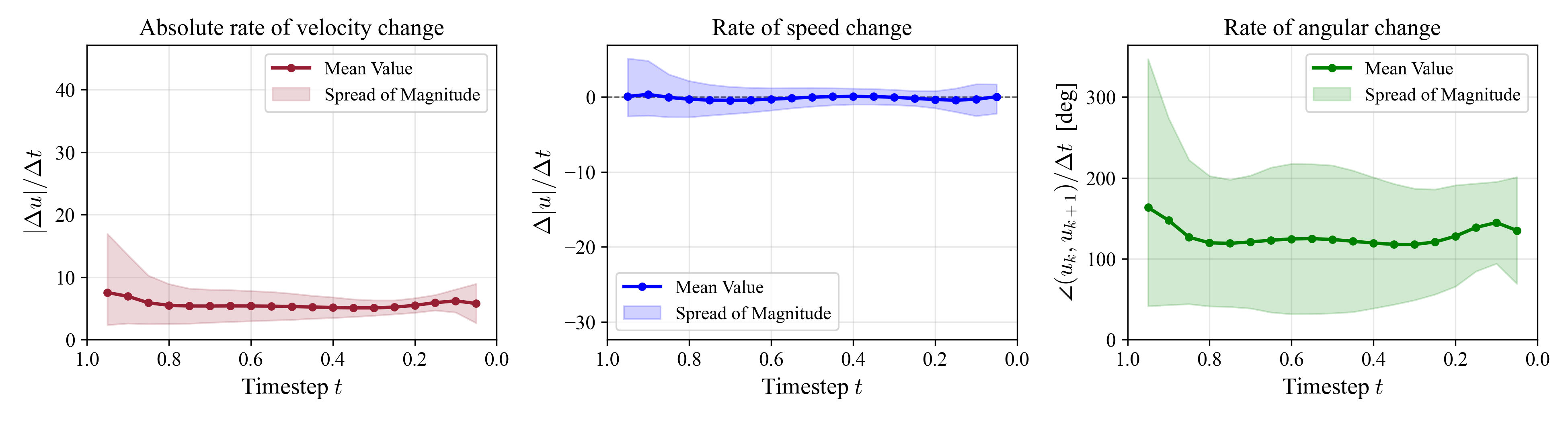}
        \caption{DiT-B on the ImageNet dataset.}
        \label{fig:mf_dynamics}
    \end{subfigure}
    
    \caption{\textbf{The absolute rate of velocity change (local acceleration), rate of speed change, and rate of angular change for models trained on LIBERO and ImageNet (top: SimVLA-S; bottom: DiT-B)}. Throughout the denoising process, we compute the average velocity over uniform timesteps, maintaining a constant interval $\Delta t =0.05$. We visualize the spread of magnitudes using the interval $[Q_1-1.5\mathrm{IQR}, Q_3+1.5\mathrm{IQR}]$, defined by the standard 1.5-IQR rule for identifying potential outliers. Here, $\mathrm{IQR}=Q_3-Q_1$ is the interquartile range, with $Q_1$ and $Q_3$ denoting the 25th and 75th percentiles, respectively. Values outside this interval are considered potential outliers and excluded from visualization. Each curve is aggregated over 4K samples (action chunks for LIBERO and images for ImageNet), consistent with the setup in Figure~\ref{fig:teaser:a} and Figure~\ref{fig:3} of the main paper.}
    \label{fig:comparison_mf_imf}
    \vspace{-3mm}
\end{figure}

\textbf{Q1: Why does K-MF sample $z_c$ along the conditional flow path?}

As discussed in Section~\ref{sec:kmf} of the main paper, our analysis suggests that the estimation error of the time derivative term contributes to the performance collapse of MeanFlow.
If the intermediate state $\mathbf{z}_c$ were obtained along the current model's estimated marginal flow from $\mathbf{z}_t$,
the subsequent derivative evaluation,
$\frac{d\mathbf{u}_{c,r}}{dc}$,
would also depend on the accuracy of this predicted state.
Errors in the first estimation
could therefore perturb the predicted intermediate state and,
in turn, affect the evaluation of
$\frac{\mathrm{d}\mathbf{u}_{c,r}}{\mathrm{d}c}$.
Particularly in the early training stage, this issue could destabilize the bootstrapping feedback loop and expose the method to similar training collapse.

Differently, we construct
$\mathbf{z}_c=(1-c)\mathbf{a}+c\boldsymbol{\epsilon}$
along the conditional flow path, directly anchoring the intermediate
state to the training data.
Since $\mathbf{z}_c$ does not depend on the model's estimation of
$\mathbf{u}_{t,c}$, the derivative
$\frac{\mathrm{d}\mathbf{u}_{c,r}}{\mathrm{d}c}$
can be evaluated without inheriting prediction
errors in $\mathbf{z}_c$ from the first estimation. This is why we refer to this operation as \textbf{decoupling}. This decoupling also enables the two derivative evaluations to be performed in parallel, since neither depends on the result of the other.
This improves training efficiency, as discussed in the main paper.

\textbf{Q2: RFMs typically require fewer sampling steps than image generation models, but why is one-step generation for RFMs more challenging?}

To answer this question, we further show speed and angular change to better compare the differences between RFMs and image generation models.
As shown in Figure \ref{fig:comparison_mf_imf}, we observe: (1) overall, the RFM shows sharp variation trends across three metrics at the late denoising stage, in contrast to the stability in image generation models (although the rate of angular change exhibits a broad spread, neither its mean nor its spread shows a sharp late-stage increase); (2) in the velocity field of RFMs, the surge of local acceleration in the late denoising stage primarily stems from the combined effects of sharp variations in both speed and angular changes.
This explains why flow matching typically requires fewer sampling steps to generate actions in RFMs than in image generation models: the velocity changes slowly in both magnitude and direction in the early denoising stage, yielding a nearly straight flow path that can be approximated over a longer time interval using the local instantaneous velocity. 
When we move to one-step generation, two issues arise.
First, the local-global inconsistency makes it difficult to estimate the time derivative term locally. Second, the resulting estimation errors tend to be amplified with the widening spread of local acceleration. The joint effect of these two issues makes it difficult to obtain an accurate time derivative, ultimately giving rise to this counterintuitive phenomenon.

Finally, our findings point to an emerging challenge for one-step action generation: as RFMs scale to larger datasets and more complex, high-DoF embodiments, the difficulties identified in this work may become more pronounced. This work takes the first step toward addressing these challenges, and we hope our observations, insights, and proposed method will inspire future research in more demanding and challenging settings.

\label{sec:analysis}

\begin{algorithm}[]
\caption{Kinematic MeanFlow (K-MF) Training}
\label{alg:kmf_train}
\small
\begin{algorithmic}[1]
\State \textbf{Input:} Training action chunk $\mathbf{a}$, policy $\pi_\theta$, generator $G_\phi$ (optional), balancing factor $\alpha$
\State Sample $t, r \sim \text{Beta}(1.5, 1.0)$ ($t > r$), \ $\boldsymbol{\epsilon} \sim \mathcal{N}(\mathbf{0}, \mathbf{I})$; \ 
\State Set $r \leftarrow t$ for 50\% of the samples \
\State $\mathbf{z}_t \leftarrow t\epsilon + (1-t)\mathbf{a}, \ 
\mathbf{v} \leftarrow \epsilon - \mathbf{a}$
\State $\mathbf{u}, \frac{\mathrm{d}\mathbf{u}_{t,r}}{\mathrm{d}t} \leftarrow \text{JVP}\big(\pi_\theta, (\mathbf{z}_t, r, t), (\mathbf{v}, 0, 1)\big)$
\State $\lambda  \sim \mathcal{U}(0, 1) \text{ (Random)} \mid \lambda \leftarrow 0.5 \text{ (Midpoint)} \mid \lambda \leftarrow  G_\phi(t, r) \text{ (Learnable)}$ 
\State $c \leftarrow r + \lambda(t-r)$, \ $\mathbf{z}_c \leftarrow c\boldsymbol{\epsilon} + (1-c)\mathbf{a}$
\State \textbf{with} \text{torch.no\_grad()}:
    \State \hspace{\algorithmicindent}$\_,
    \left[\frac{\mathrm{d}\mathbf{u}_{t,c}}{\mathrm{d}t},
    \frac{\mathrm{d}\mathbf{u}_{c,r}}{\mathrm{d}c}\right]
    \leftarrow \text{JVP}\Big(
    \pi_\theta,
    \big([\mathbf{z}_t, \mathbf{z}_c], [c, r], [t, c]\big),
    \big([\mathbf{v}, \mathbf{v}], [0, 0], [1, 1]\big)
    \Big)$\
\State $\mathbf{u}_{\text{K-MF}} \leftarrow \mathbf{v} - (t - r)\left[(1-\lambda)^2 \operatorname{sg}\left(\frac{\mathrm{d}\mathbf{u}_{t,c}}{\mathrm{d}t}\right) + \lambda^2 \operatorname{sg}\left(\frac{\mathrm{d}\mathbf{u}_{c,r}}{\mathrm{d}c}\right)\right]$ 
\State $\mathcal{L}_{\text{K-MF}} \leftarrow \|\mathbf{u} - \operatorname{sg}(\mathbf{u}_{\text{K-MF}})\|_2^2$  \Comment{Ensure only $\theta$ is updated, as in Eq.~\ref{eq:kmf_loss}}
\If{Learnable}
    \State $\mathcal{L}_{\text{gen}} \leftarrow \omega \|\operatorname{sg}(\mathbf{u}) - \mathbf{u}_{\text{K-MF}}\|_2^2$ \Comment{Ensure only $\phi$ is updated, as in Eq.~\ref{eq:generator_loss}}
    \State $\mathcal{L} \leftarrow \mathcal{L}_{\text{K-MF}} + \alpha \mathcal{L}_{\text{gen}}$ 
\Else
    \State $\mathcal{L} \leftarrow \mathcal{L}_{\text{K-MF}}$ 
\EndIf
\State \textbf{Backpropagate} $\mathcal{L}$ once to obtain $\nabla_\theta \mathcal{L}_{\text{K-MF}}$ (and $\alpha\nabla_\phi \mathcal{L}_{\text{gen}}$ if Learnable).
\State \textbf{Update} $\theta$ using $\nabla_\theta \mathcal{L}_{\text{K-MF}}$ (and $\phi$ using $\alpha\nabla_\phi \mathcal{L}_{\text{gen}}$ if Learnable).
\end{algorithmic}
\end{algorithm}


\begin{algorithm}[]
\caption{Kinematic MeanFlow (K-MF) Inference}
\label{alg:kmf_infer}
\small
\begin{algorithmic}[1]
\State \textbf{Input:} policy $\pi_\theta$ 
\State Get conditions $(\mathbf{o}, \mathbf{s})$, Sample $\boldsymbol{\epsilon} \sim \mathcal{N}(\mathbf{0}, \mathbf{I})$
\State $\mathbf{a} \leftarrow \boldsymbol{\epsilon} - \pi_\theta(\boldsymbol{\epsilon}, 0, 1 \mid \mathbf{o}, \mathbf{s})$
\State \textbf{return} action chunk $\mathbf{a}$
\end{algorithmic}
\end{algorithm}

\end{document}